\documentclass[11pt, journal]{article}
\usepackage{graphicx}
\usepackage{amsmath,amssymb,amsfonts}
\usepackage{url}
\usepackage{subfigure}
\begin{document}

\title{Unsupervised Deep Learning Algorithm for PDE-based Forward and Inverse Problems}
\author{Leah Bar\footnote{barleah.libra@gmail.com}~~and Nir Sochen}
\date{Department of Applied Mathematics\\Tel-Aviv University \\Tel-Aviv 69978, Israel
\\[\baselineskip]}
\maketitle

\begin{abstract}
We propose a neural network-based algorithm for solving forward and inverse problems for partial differential equations in unsupervised fashion. 
The solution is approximated by a deep neural network which is the minimizer of a cost function, and satisfies the PDE, boundary conditions, and additional regularizations. The method is mesh free and can be easily applied to an arbitrary regular domain. 
We focus on 2D second order elliptical system with non-constant coefficients, with application to Electrical Impedance Tomography.

\end{abstract}

\section{Introduction}
Inverse problems in partial differential equations are fundamental in science and mathematics with wide applications in medical imaging, signal processing, computer vision, remote sensing, electromagnetism and more. Classical methods such as finite differences, finite volume and finite elements are numerical discretization-based methods where the domain is divided into a uniform grid or polygon mesh. The differential equation is then reduced to a system of algebraic equations. These methods may have some limitations: the solution is numeric and may suffer from high condition number, highly dependent on the discretization and even the second derivative is sensitive to noise.

In the last few years, deep learning and neural network-based algorithms are extensively used in pattern recognition, image processing, computer vision and more. Recently, the deep learning approach had been adopted to the field of PDEs as well by converting the problem into a machine learning one. 
In \emph{Supervised learning}, the network maps an input to an output based on example input-output pairs. This strategy  
is used in inverse problems, where the input to the network is a set of observations/measurements (e.g. x-ray tomography, ultrasound) and the output is the set of parameters of interest (tissue density etc.) \cite{Feig18,Molina18,Unser17}.
\emph{Unsupervised learning} on the other hand is a self-learning mechanism where the natural structure presents within a set of data points is inferred.  

Algorithms for forward and inverse problems in partial differential equations via  unsupervised learning were recently introduced. The \emph{indirect} approach utilizes a neural network as a component in the solution. Li et al.~\cite{Li18} for example, proposed the NETT (Network Tikhonov) approach to inverse problems. NETT considers
regularized solutions having small value of a regularizer defined by a trained neural network. Khoo and Ying~\cite{Khoo18} introduced a novel neural network architecture, SwitchNet, for solving the wave
equation based inverse scattering problems via providing maps between the scatterers and the scattered field. Han et al.~\cite{Han18} developed a deep learning-based
approach that can handle general high-dimensional parabolic PDEs. To this end, the PDEs 
are reformulated using backward stochastic differential equations.
The latter is solved by a temporal discretization and the gradient of the
unknown solution at each time step is approximated by neural network. 

\emph{Direct} algorithms solve the forward problem PDEs by directly approximating the solution with a deep neural network. The network parameters are determined by the optimization of a cost function such that the optimal solution satisfies the PDE, boundary conditions and initial conditions.
Chiaramonte and Kiener~\cite{chiara} addressed the forward problem by constructing a one layer network which satisfies the PDE within the domain. The boundary conditions were analytically integrated in the cost function. They demonstrated their algorithm on the Laplace and hyperbolic conservation law PDEs.
Sirignano and Spiliopoulos~\cite{sir17} proposed a deep learning
forward problem solver for high dimensional PDEs. Their algorithm was demonstrated on the American option free-boundary equation. 
Raissi et al.~\cite{Raissi17} focused on continuous time models and solved the Burgers and Shr\"{o}dinger equations.

In this work we focus on the forward and inverse PDEs problems via a direct unsupervised method. Our key contributions are three fold: (1) in the forward part we extend the standard $L_2$-based fidelity term in the cost function by adding $L_\infty$-like norm. Moreover, (2) some regularization terms which impose a-priori knowledge on the solution can be easily incorporated. (3) An important feature of our construction is the ability to handle free-form domain in a mesh free manner. We demonstrate our algorithm by a second order elliptic equation, in particular the Electrical Impedance Tomography (EIT) application.

\section{Mathematical Formulation}
Let $\Omega$ be a bounded open and connected subset of $\mathbb{R}^d$, and $A = A(x) = (a^{ij} (x))$ be any given $d \times d$ symmetric positive definite matrix of functions for $1 \leq i,j \leq d$. Let $b = b(x) = (b^j(x))$ be any
given n-tuple of functions and let $c = c(x)$ be any given function.
A second order operator $\mathcal{L}$ is said to be in divergence form,
if $\mathcal{L}$ acting on some $u$ has the form
\begin{equation}
\mathcal{L}u = \partial_i(a^{ij}(x)\partial_ju)+b^j(x)\partial_ju+c(x)u,~~i,j=1,\ldots,
\label{eq:ellipgen}
\end{equation}
where we use the Einstein summation convention.
Consider the partial differential problem with Dirichlet boundary conditions
\begin{equation}
\begin{aligned}
&\mathcal{L}u = 0,&x&\in\Omega\\
&u(x) = u_0(x),&x&\in\partial\Omega.
\end{aligned}
\end{equation}
The \emph{forward problem} solves $u$ given the coefficients $\theta:=\{a^{ij}(x),b^j(x),c(x)\}$ while the \emph{inverse problem} determines the coefficients set $\theta$ given $u$.

\begin{figure}
\begin{center}
\includegraphics[width=.95\linewidth]{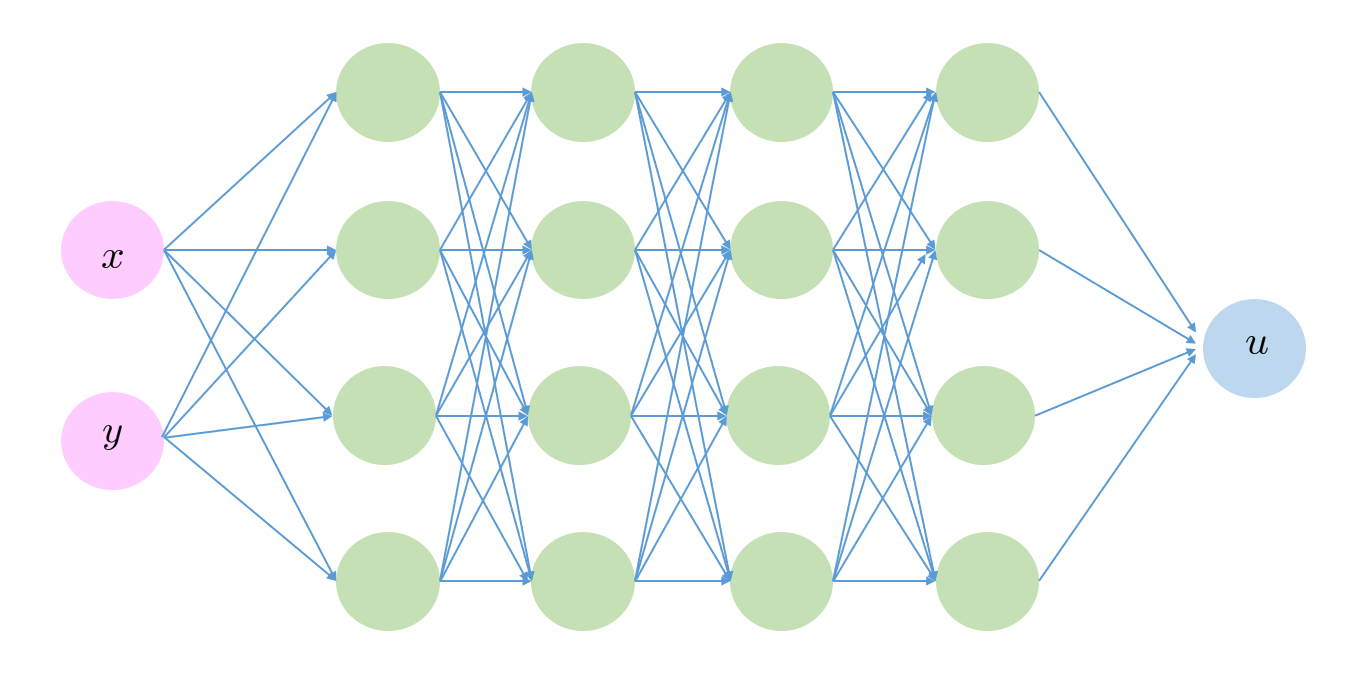}
\end{center}
\caption{Network architecture: the point $(x,y)\in \mathbb{R}^2$ serves as an input and $u$ as the output.}
\label{fig:net}
\end{figure}

The proposed algorithm approximates the solutions in both problems by neural networks $u(x;w_u),a^{ij}(x;w_{ij}),\ldots,c(x;w_c)$ such that the networks are parameterized by $w_u,w_{ij},\ldots,w_c$,
and the input to the network is $x\in \mathbb{R}^d$. Figure~\ref{fig:net} depicts a network architecture of $u$ in $\mathbb{R}^2$.
The network consists of few fully connected layers with \emph{tanh} activation and linear sum in the last layer.

The network is trained to satisfy the PDE with the boundary conditions by minimizing a cost function. In the forward problem
\begin{equation}
\mathcal{F}(u) = \lambda\|\mathcal{L}u\|_{2}^2+\mu\|\mathcal{L}u\|_{\infty}+\|u-u_0\|_{1,\partial\Omega}+\mathcal{R}^F(u),
\label{eq:Fu}
\end{equation}
and in the inverse problem
\begin{equation}
\mathcal{I}(a^{ij}) = \lambda\|\mathcal{L}u\|_{2}^2+\mu\|\mathcal{L}u\|_{\infty}+\|a^{ij}-a^{ij}_0\|_{1,\partial\Omega}+\mathcal{R}^I(a^{ij}).
\label{eq:Fi}
\end{equation}
The first two terms enforce the solution to satisfy the equation. The first term minimizes the error in $L_2$ sense while the second term minimizes the maximal error. 
This term is important since the $L_2$ term only forces the equation up to a set of zero measure. The $L_\infty$ term takes care of possible outliers. 
The third term imposes boundary conditions and the last term is a regularizer which can be tailored to the application.   
There are few advantages of this setting. First, the solutions are smooth analytic functions and are therefore \emph{analytically differentiable}. In addition, this framework enables setting of a prior knowledge on the solution by designing the regularizers $\mathcal{R}^F$ and $\mathcal{R}^I$. 
Lastly, the training procedure is mesh free. In the sequel, we use \emph{random} points in the domain and its boundary in the course of the optimization of~\eqref{eq:Fu} and ~\eqref{eq:Fi}. This means that  the solution does not depend upon a coordinate mesh and we can also define in principle an arbitrary regular domain $\Omega$. 
\section{Application to Electrical Impedance Tomography}
Let us address a special case of~\eqref{eq:ellipgen},
\begin{equation}
\begin{aligned}
&\nabla \cdot\Big( \sigma(x) \nabla u(x)\Big)=0,~~&x&\in\Omega\subset \mathbb{R}^2\\
&u(x)=u_0(x),&x&\in\partial \Omega.
\end{aligned}
\label{eq:ellip}
\end{equation}
We assume that $0< \sigma(x) \in C^1(\Omega)$, which guarantees existence and uniqueness of a solution $u \in C^2(\Omega)$~\cite{Evans}.

The elliptical system~\eqref{eq:ellip} was addressed by Siltanen et al.~\cite{Siltanen_2000} in the context of Electrical Impedance Tomography (EIT) which is a reconstruction method for the inverse conductivity problem. The function $\sigma$ stands for the electrical conductivity density, and $u$ is the electrical potential. An electrical current
$$ 
\psi_n = \sigma\frac{\partial u_n}{\partial \nu}\Big|_{\partial\Omega} = \frac{1}{\sqrt{2\pi}}\cos{n\varphi},~~n\in\mathbb{Z}
$$ 
is applied on electrodes on the surface $\partial\Omega$, where $\varphi$ is the angle in polar coordinate system along the domain boundary and $\nu$ is the normal unit. The resulting voltage $u|_{\partial\Omega}=u_0$ is measured through the electrodes. 
The conductivity $\sigma$ is determined from the knowledge of the Dirichlet-to-Neumann map or voltage-to-current map
$$
\Lambda_\gamma:u|_{\partial\Omega}\to\sigma\frac{\partial u_n}{\partial \nu}\Big|_{\partial_\Omega}
$$ 
using the D-bar method~\cite{Siltanen_2012}.

We demonstrate our framework by solving the forward and inverse problem of~\eqref{eq:ellip} which is a first step towards a full tomography.
Following Mueller and Siltanen~\cite{Siltanen_2012}, we simulate the voltage measurement $u|_{\partial\Omega}$ by the Finite Element Method (FEM) given two variants of a conductivity phantom $\sigma(x)$ on the unit disc.  We calculate the FEM solution with different triangle mesh densities such that finer meshes do not improve the numerical solution.

\begin{figure}
\begin{center}
\includegraphics[width=4cm]{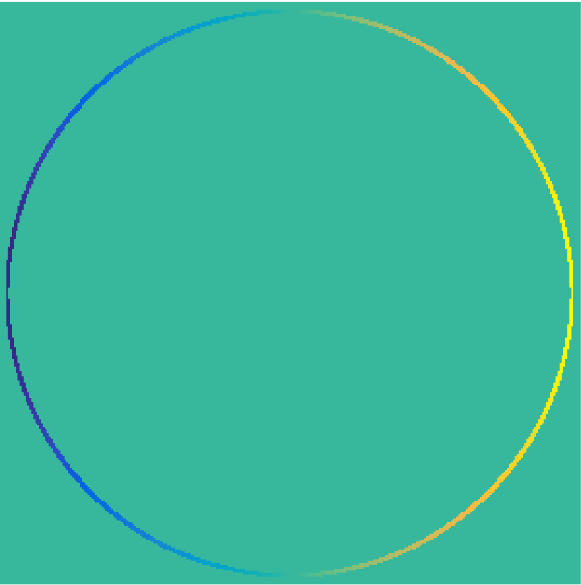}
\includegraphics[width=4cm]{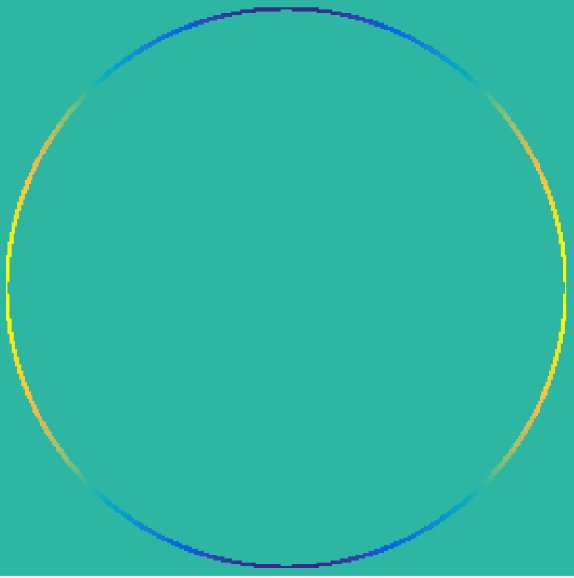}
\includegraphics[width=4cm]{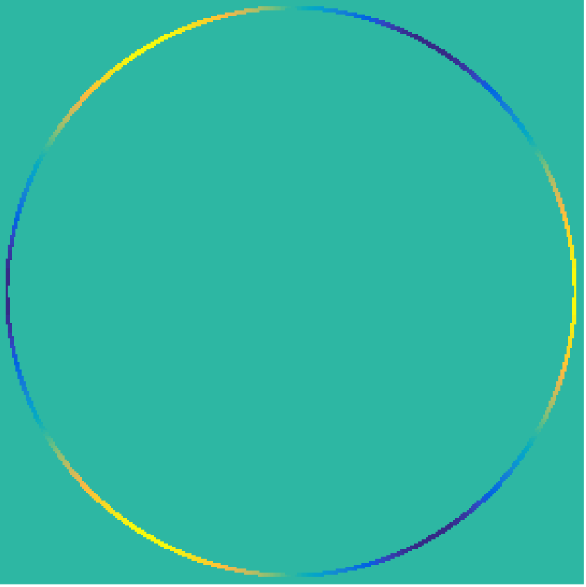}
\end{center}
\caption{Electrical current $\psi_n$ for $n=1,2,3$}
\label{fig:current}
\end{figure}

With our suggested method, the forward problem determines the electrical potential $u$ in the whole domain $\Omega$, while the inverse problem uses the approximated $u$ and calculates the conductivity $\sigma$ given that $\sigma|_{\partial\Omega}=\sigma_0$. Throughout the paper we use three different electrical currents $\psi_n$ where $n=1,2,3$, see Figure~\ref{fig:current}.

\section{Forward Problem}
In the forward problem the conductivity $\sigma(x_i)$ and boundary conditions $u_0(x_b)$ are given for random points set $\{x_i\}\in\Omega\subset\mathbb{R}^2$, $\{x_b\}\in\partial\Omega\subset\mathbb{R}^2$ with sets size of $N_s$ and $N_b$ respectively.  
A neural network having the architecture shown in Figure~\ref{fig:net} approximates $u(x)$. 
Let 
\begin{equation}
\mathcal{L}_{i}:=\nabla\cdot \Big(\sigma(x_i) \nabla u(x_i)\Big).
\label{eq:Li}
\end{equation}
The cost function~\eqref{eq:Fu} is then rewritten as
\begin{equation}
\begin{aligned}
\mathcal{F}(u(x;w_u) = &\frac{\lambda}{N_s}\sum_{i=1}^{N_s} |\mathcal{L}_{i}|^2+
\frac{\mu}{K}\sum_{k\in \text{top}_K(|\mathcal{L}_i|)}|\mathcal{L}_k|\\+
&\frac{1}{N_b}\sum_{b=1}^{N_b}\Big|u(x_b)-u_0(x_b)\Big|
+\alpha\|w_u\|_2^2.
\end{aligned}
\end{equation}
The first term is the $L_2$ norm of the differential operator, the second term is a relaxed version of the infinity norm where we take the mean value of the top-K values of $|\mathcal{L}_i|$. The third term imposes the boundary conditions and the last term serves as a regularizer of the network parameters.

The network was trained with $4$ layers having $26,26,26$, and $10$ neurons. 
The algorithm was implemented by TensorFlow~\cite{tensorflow2015} using the ADAM optimizer which is a variant of the SGD algorithm. We used batch size=$1000$ and a decaying learning rate starting at $(1e-3,1e-2,5e-4)$ corresponding to $n=1,2,3$. The learning rate was factored by $0.8$ every $200$ epochs. The algorithm parameters were set to $Ns=45000$, $N_b=1200$, $\lambda=0.01$, $\alpha=1e-8$, $K=40$ and $\mu=1e-2$.

\begin{figure}
\begin{center}
\includegraphics[width=.6\linewidth]{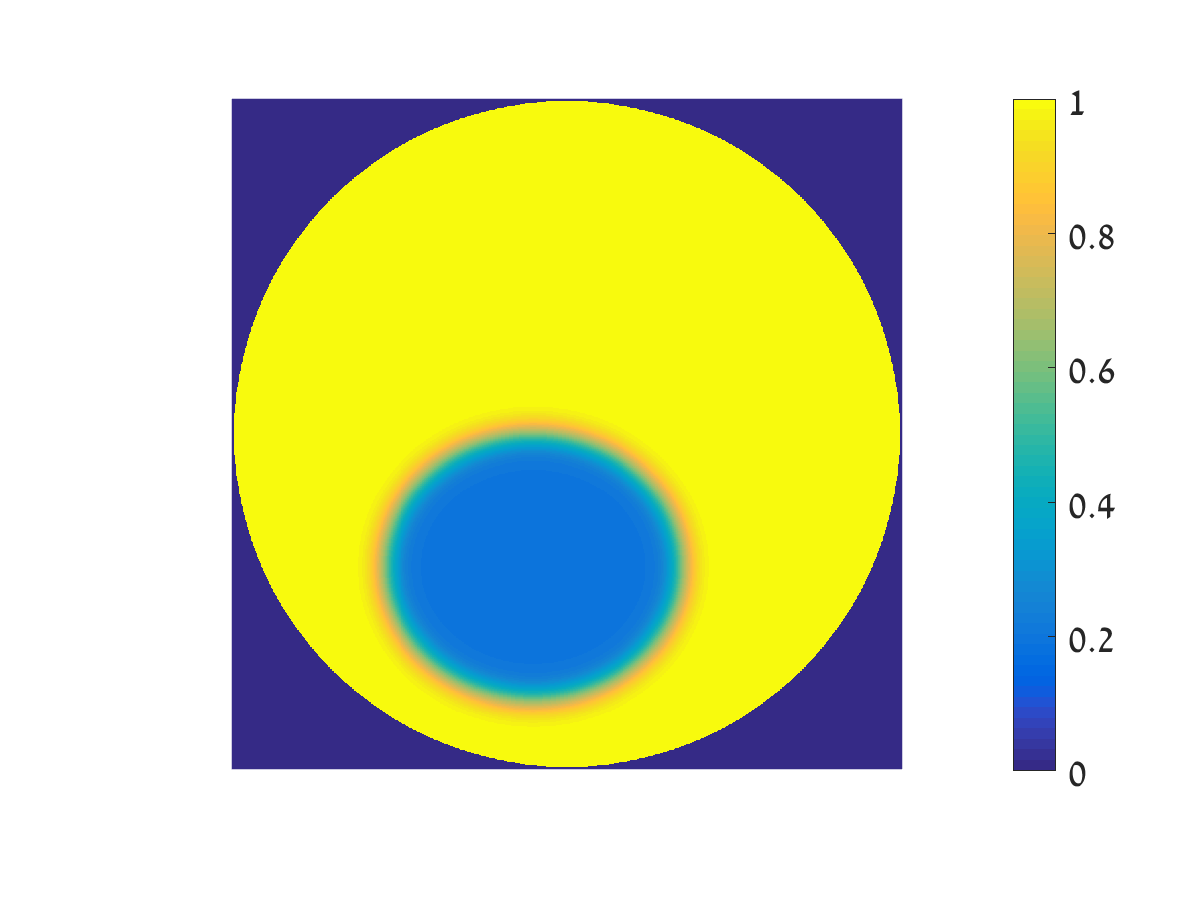}
\end{center}
\caption{The conductivity $\sigma$ of phantom $1$}
\label{fig:sig1}
\end{figure}

The first phantom is shown in Figure~\ref{fig:sig1}. The background has conductivity $1$ and the circle has conductivity $0.2$. 
The original piecewise constant function $\sigma$ was slightly smoothed by a Gaussian kernel.
 
\begin{figure}
\begin{center}
\subfigure[FEM, $n=1$]{\includegraphics[width=0.32\linewidth]{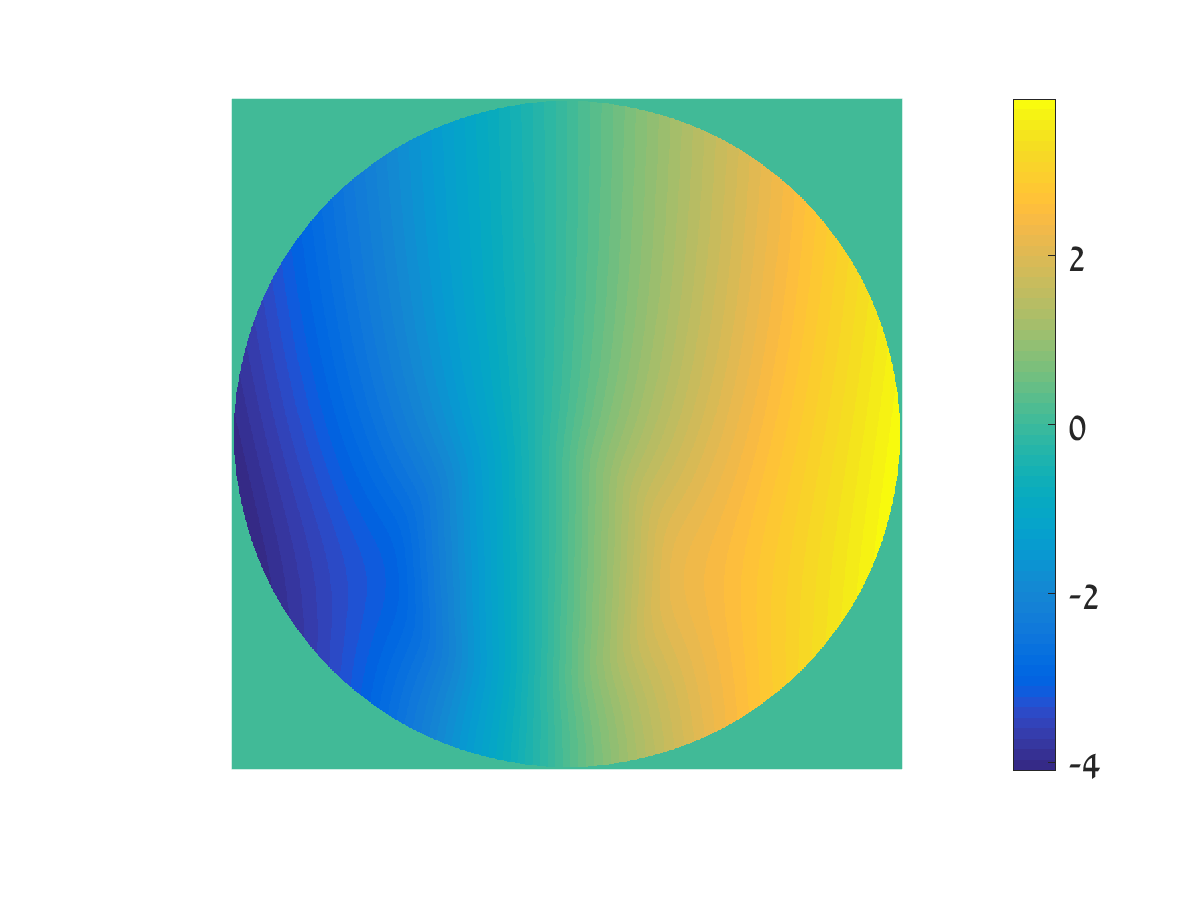}}
\subfigure[FEM, $n=2$]{\includegraphics[width=0.32\linewidth]{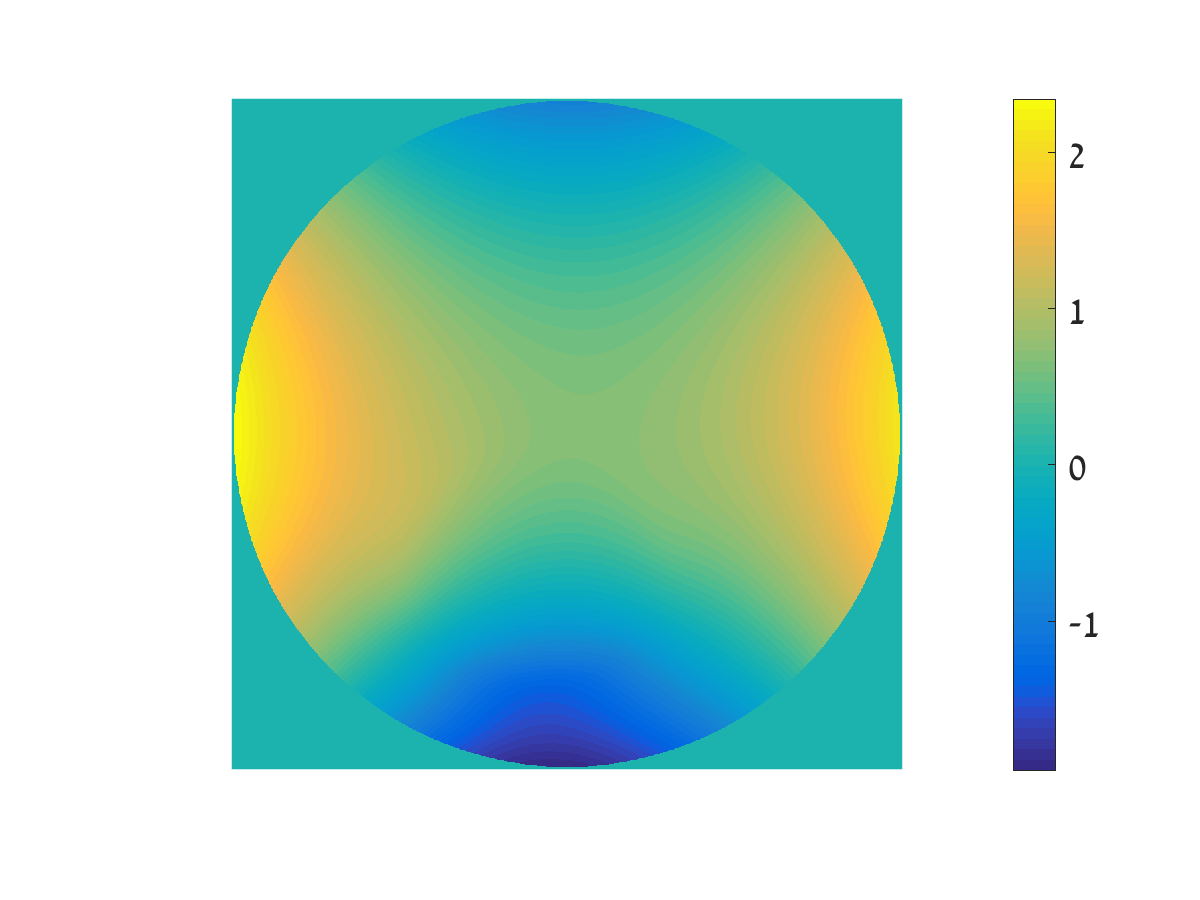}}
\subfigure[FEM, $n=3$]{\includegraphics[width=0.32\linewidth]{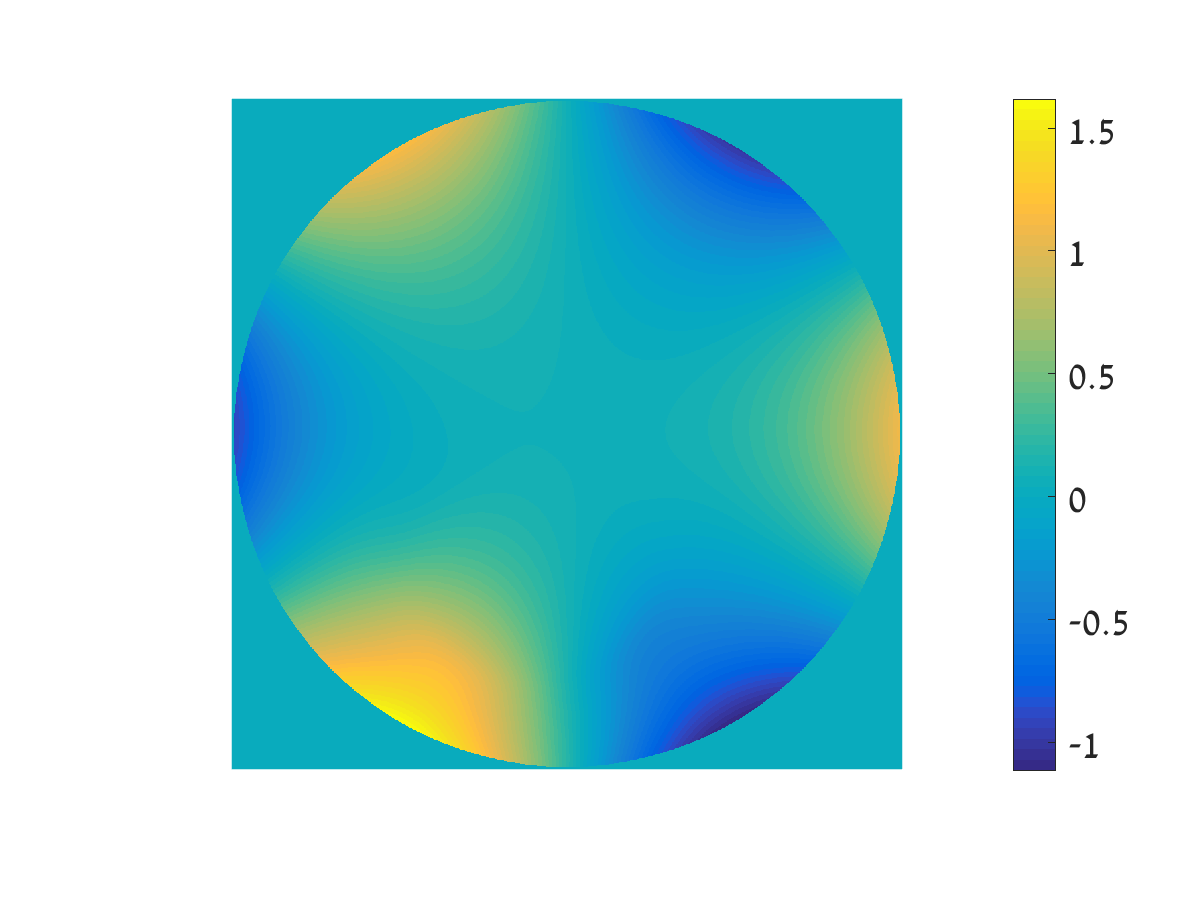}}\\[0.2cm]
\subfigure[Proposed, $n=1$]{\includegraphics[width=.32\linewidth]{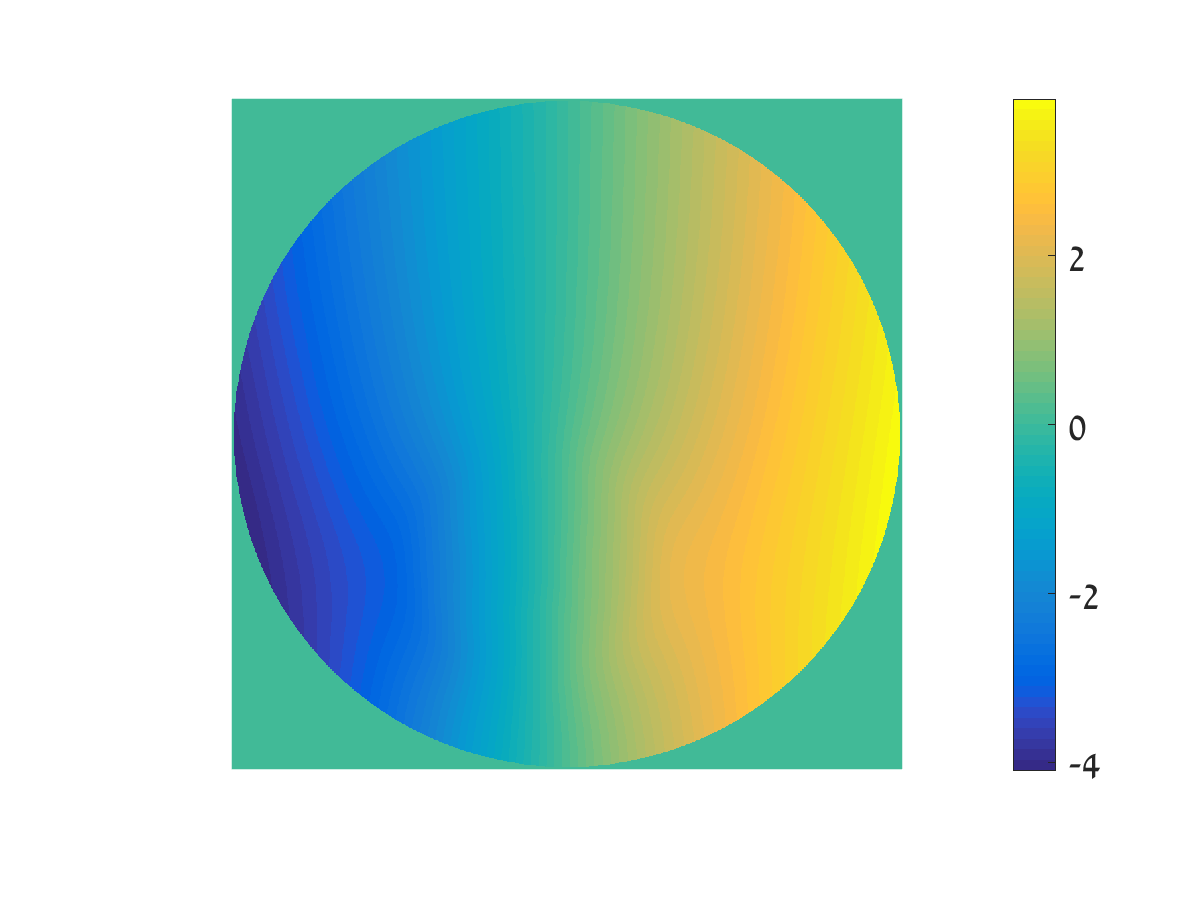}}
\subfigure[Proposed, $n=2$]{\includegraphics[width=.32\linewidth]{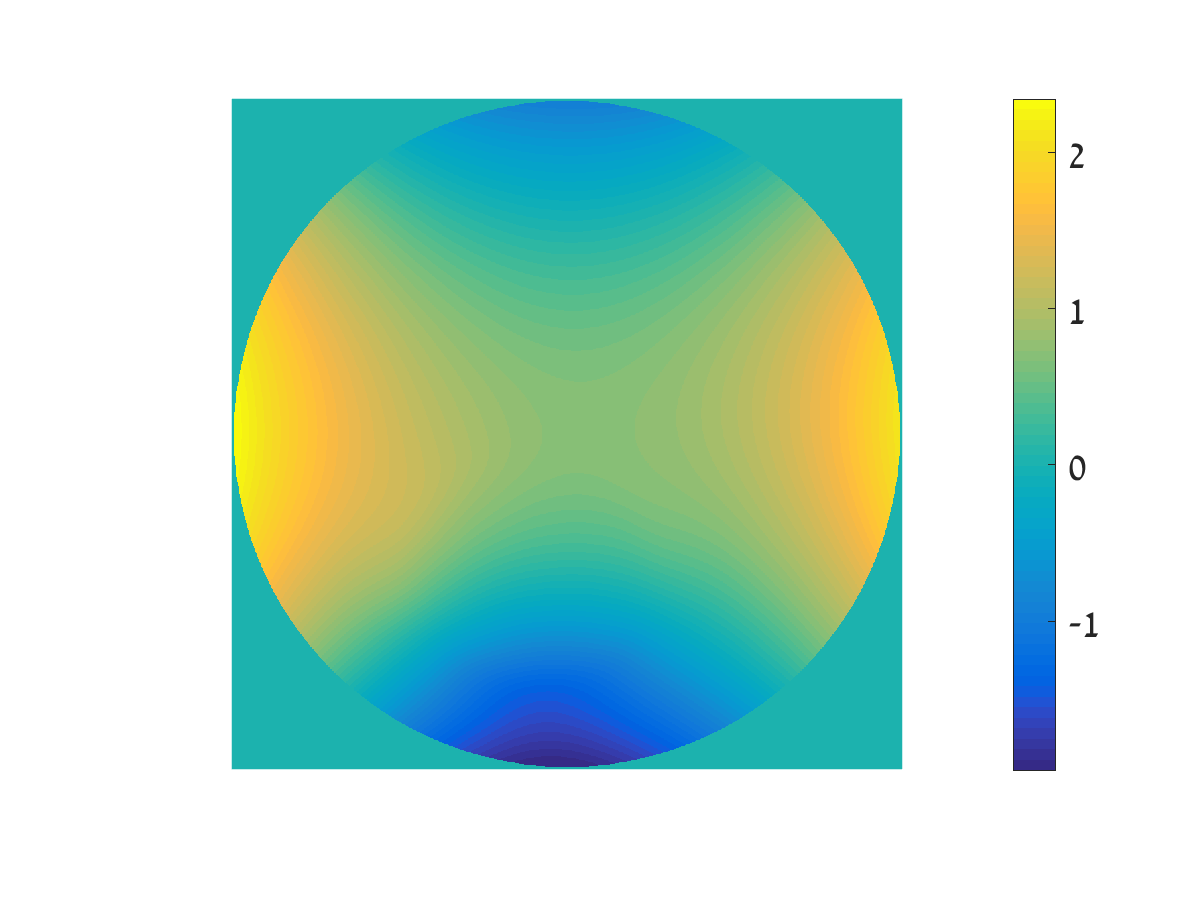}}
\subfigure[Proposed, $n=3$]{\includegraphics[width=.32\linewidth]{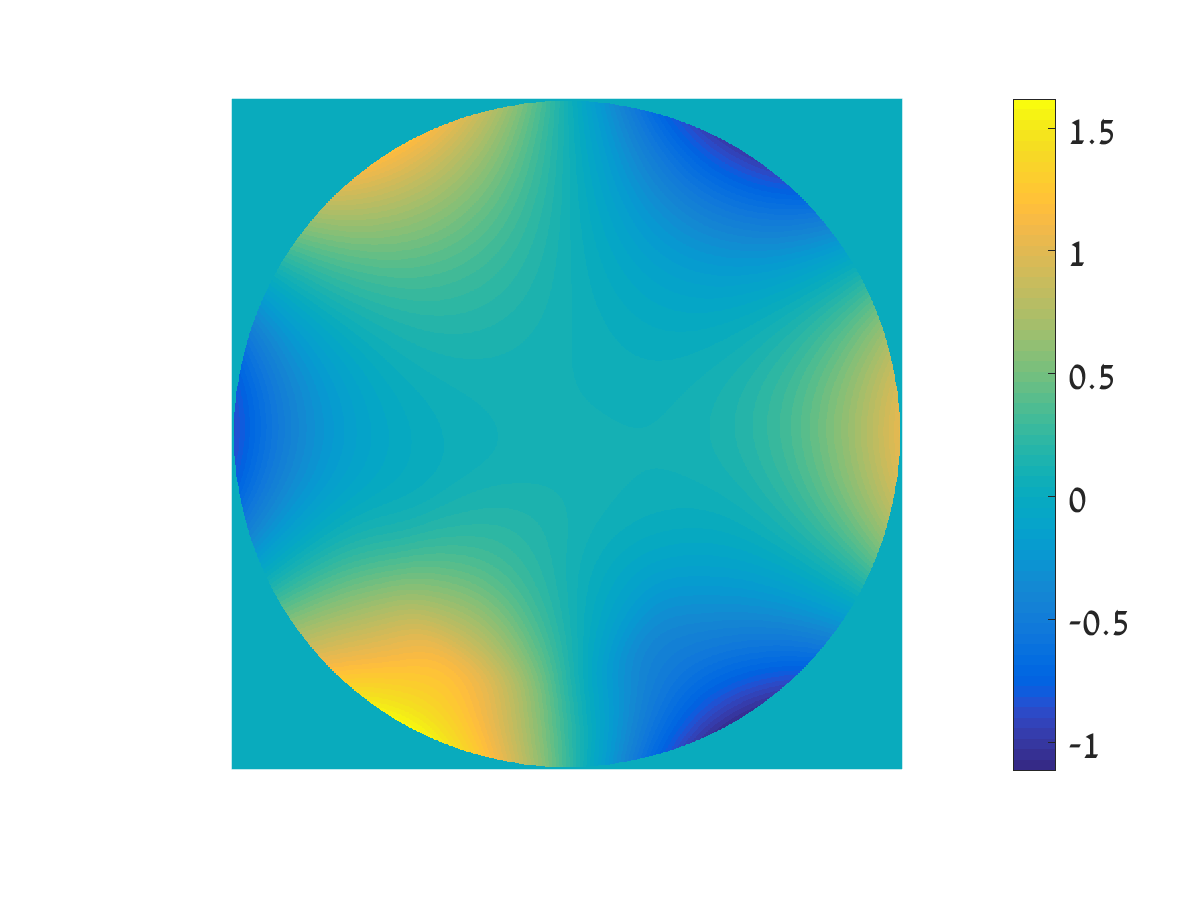}}\\[0.2cm]
\subfigure[Error, $n=1$]{\includegraphics[width=.32\linewidth]{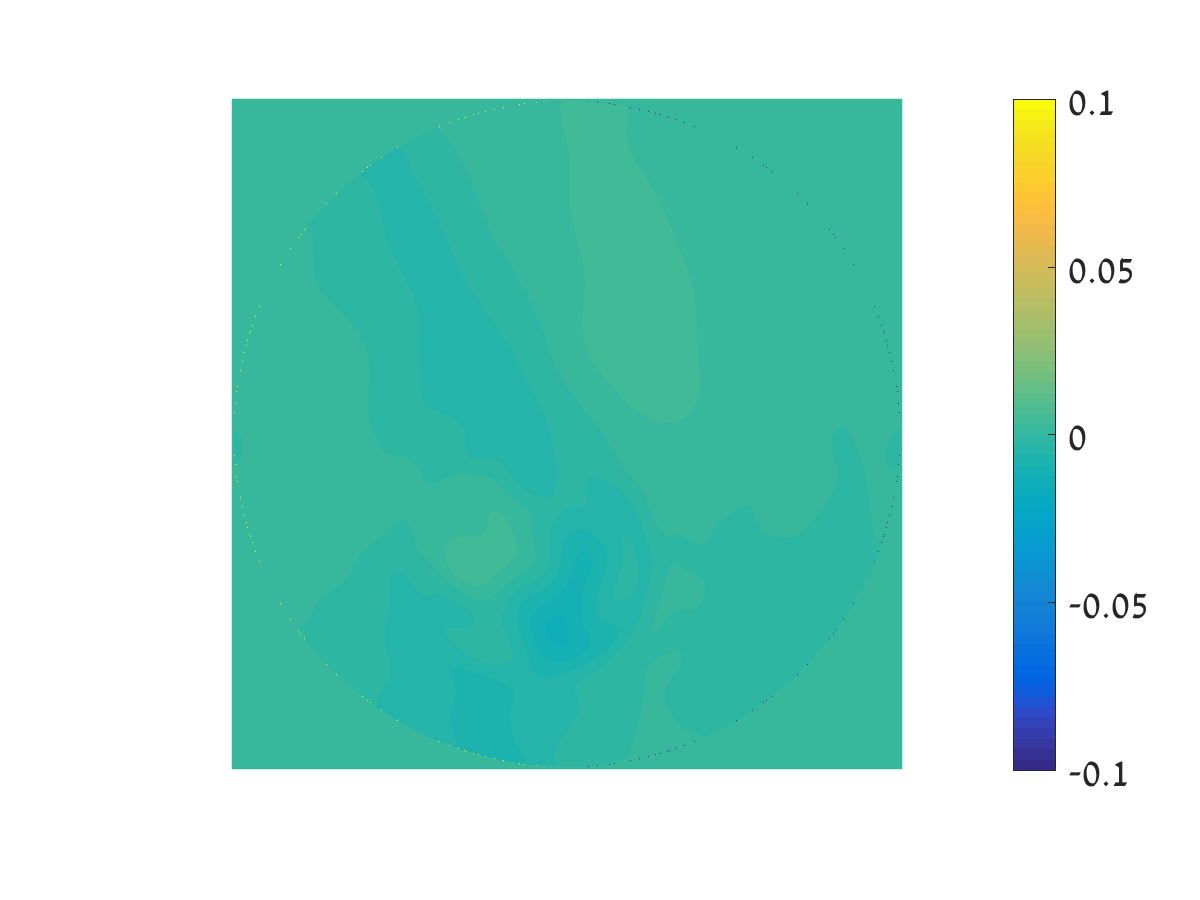}}
\subfigure[Error, $n=2$]{\includegraphics[width=.32\linewidth]{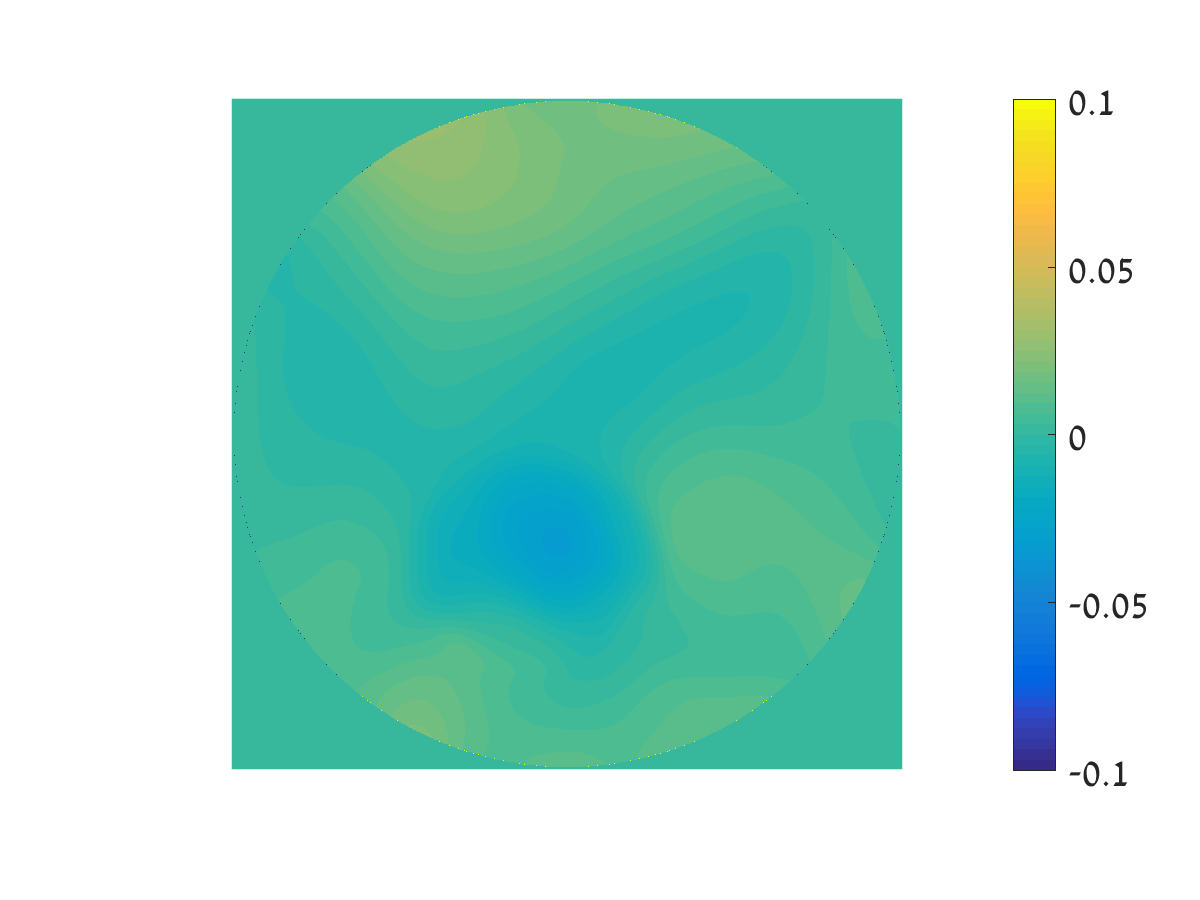}}
\subfigure[Error, $n=3$]{\includegraphics[width=.32\linewidth]{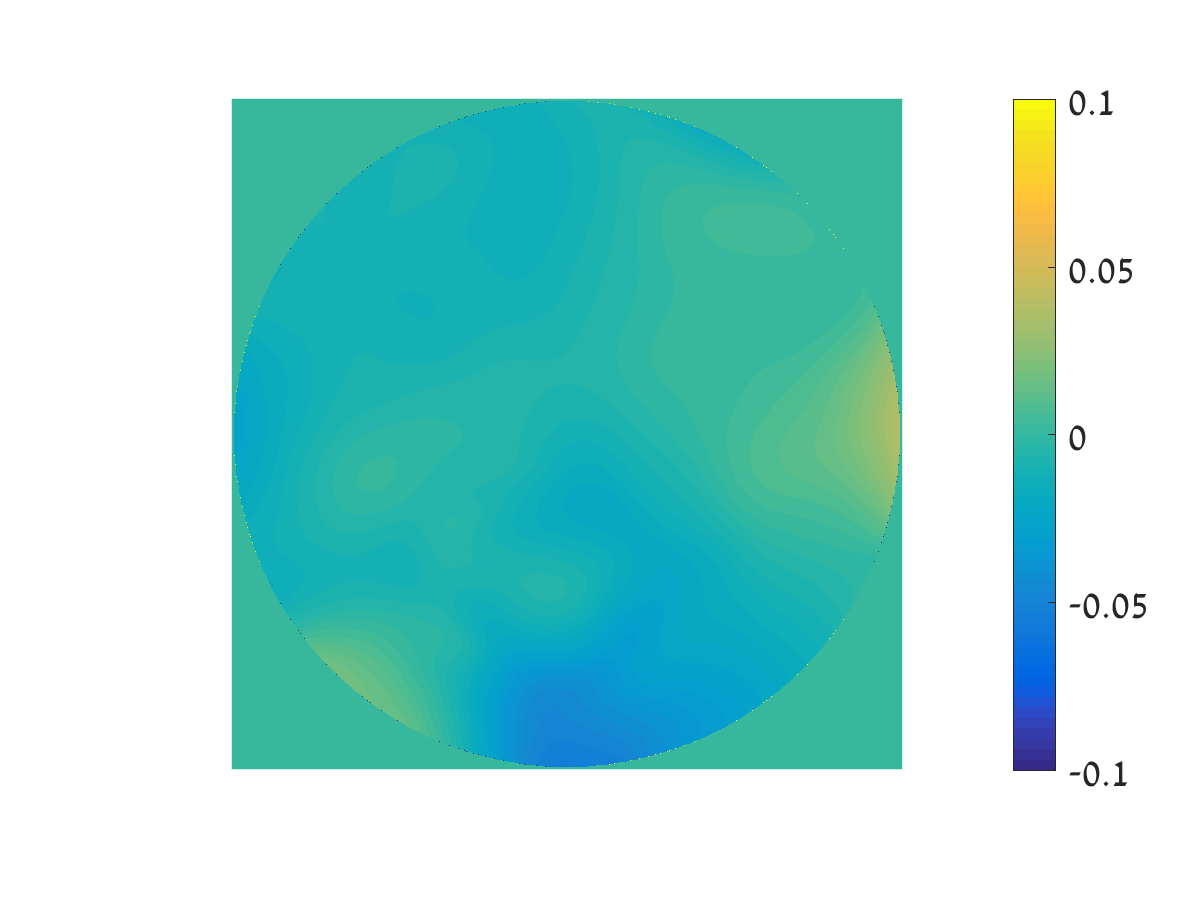}}
\end{center}
\caption{Top: ground truth (FEM) of $u$ for $n=1,2,3$ given phantom $1$. Middle: reconstruction by the proposed method. MSE = $(3.15e-3,1.33e-3,6.93e-4)$, PSNR=$(37.26,36.12,35.76)$. Bottom: relative error }
\label{fig:u1_res}
\end{figure}

Figure~\ref{fig:u1_res} summarizes the forward problem results for currents $\psi_1,\psi_2$ and $\psi_3$. The top row is the FEM solution which is referred to as ground truth. The middle row depicts the outcome of the trained network, and the bottom row shows the relative error $e$,
$$
e(x,y) = \frac{u_{\text{fem}}(x,y)-u(x,y)}{\max(u_{\text{fem}})}.
$$
Mean square errors and PSNR are indicated in the figures' caption.

\begin{figure}
\begin{center}
\subfigure[FEM, $n=1$]{\includegraphics[width=.32\linewidth]{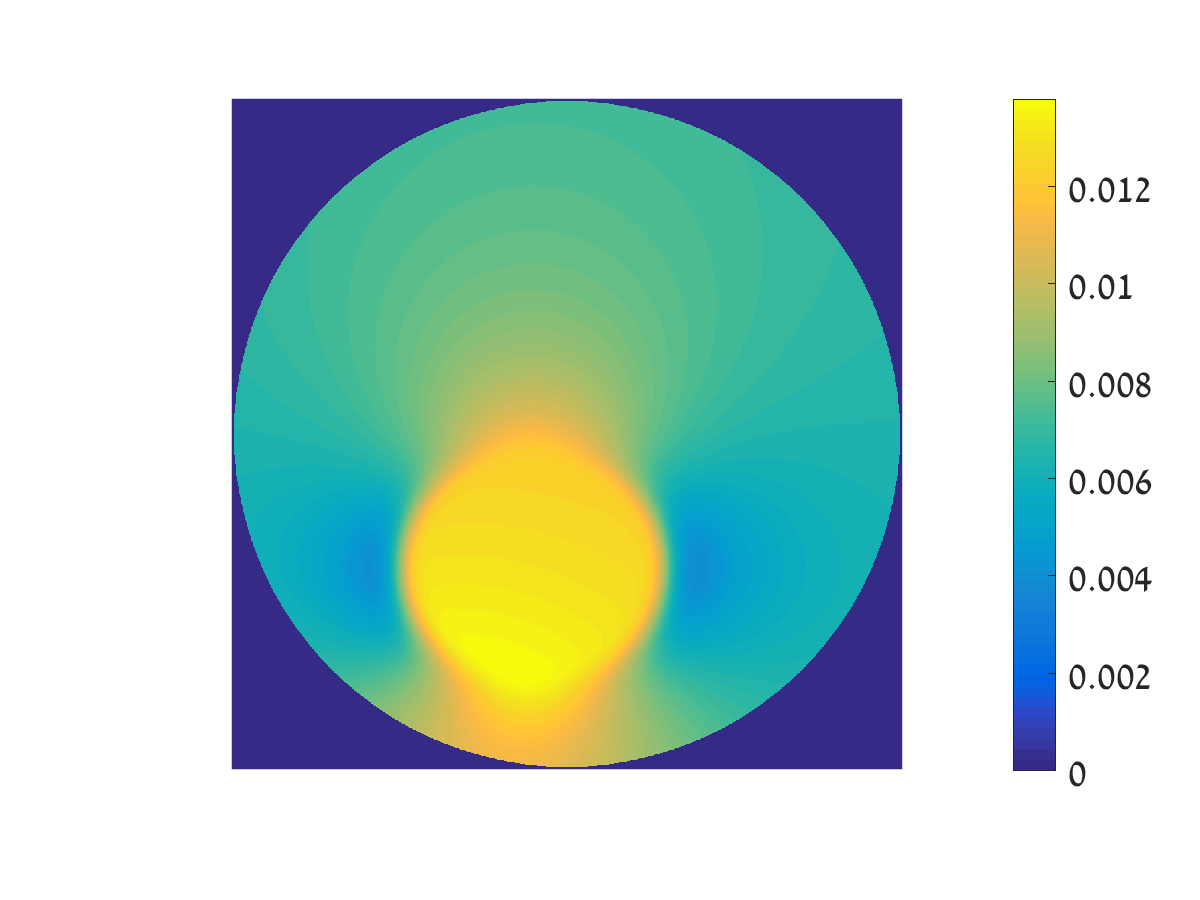}}
\subfigure[FEM, $n=2$]{\includegraphics[width=.32\linewidth]{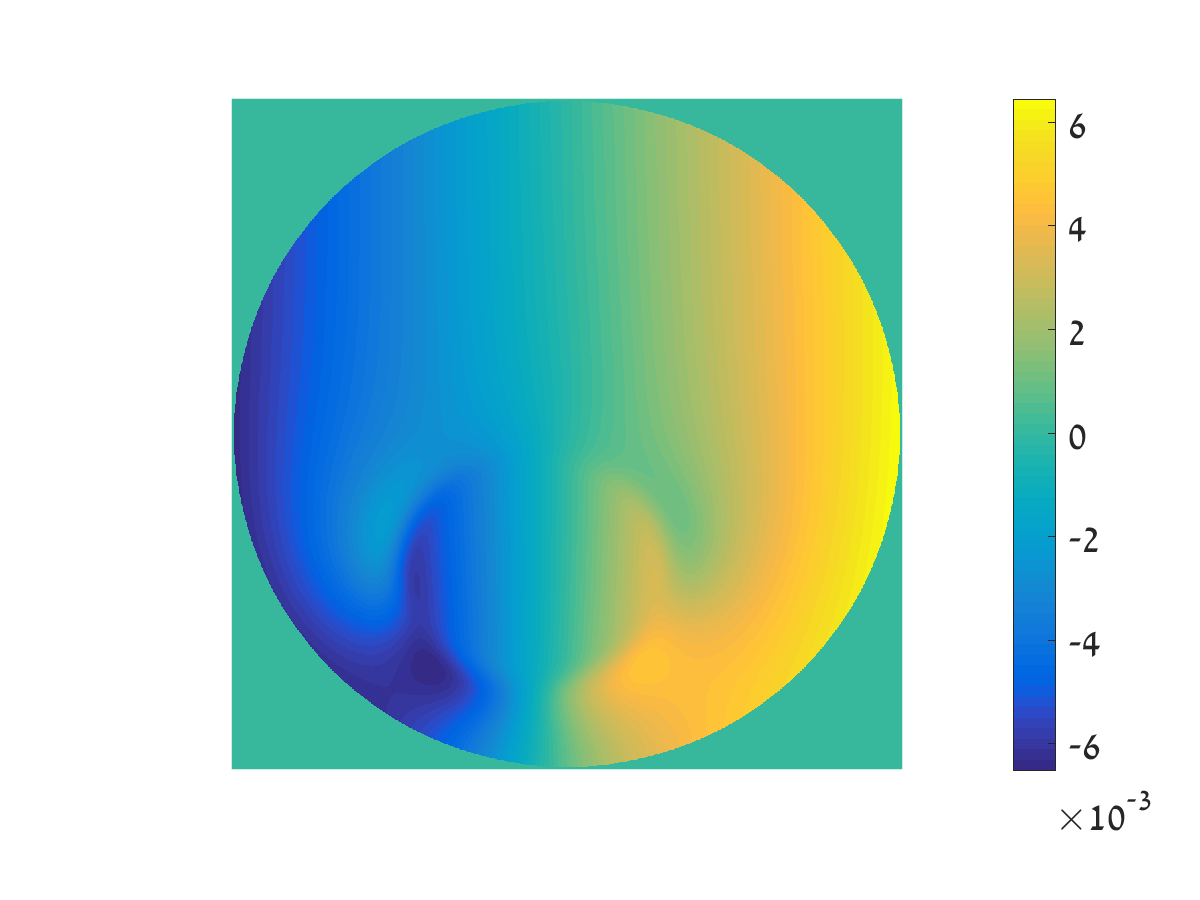}}
\subfigure[FEM, $n=3$]{\includegraphics[width=.32\linewidth]{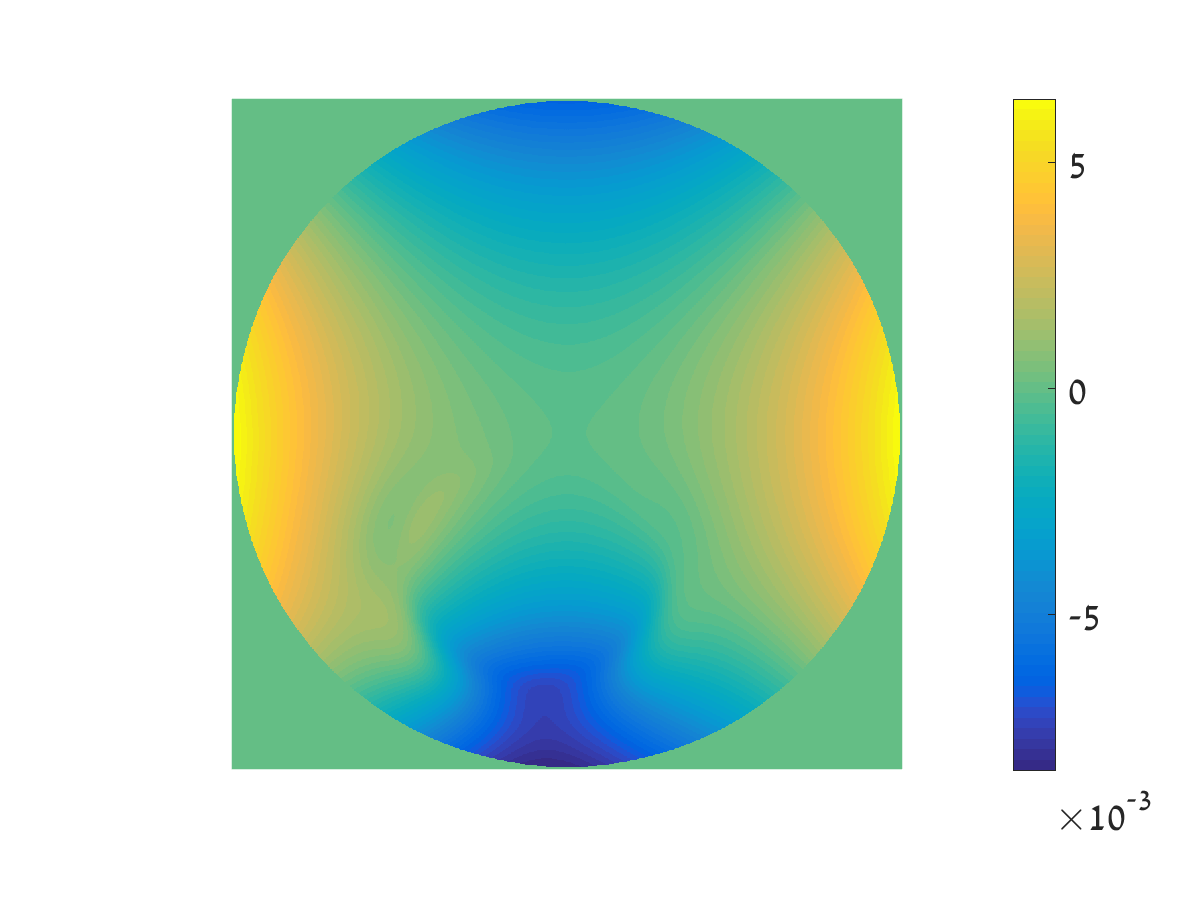}}\\[0.2cm]
\subfigure[Proposed, $n=1$]{\includegraphics[width=.32\linewidth]{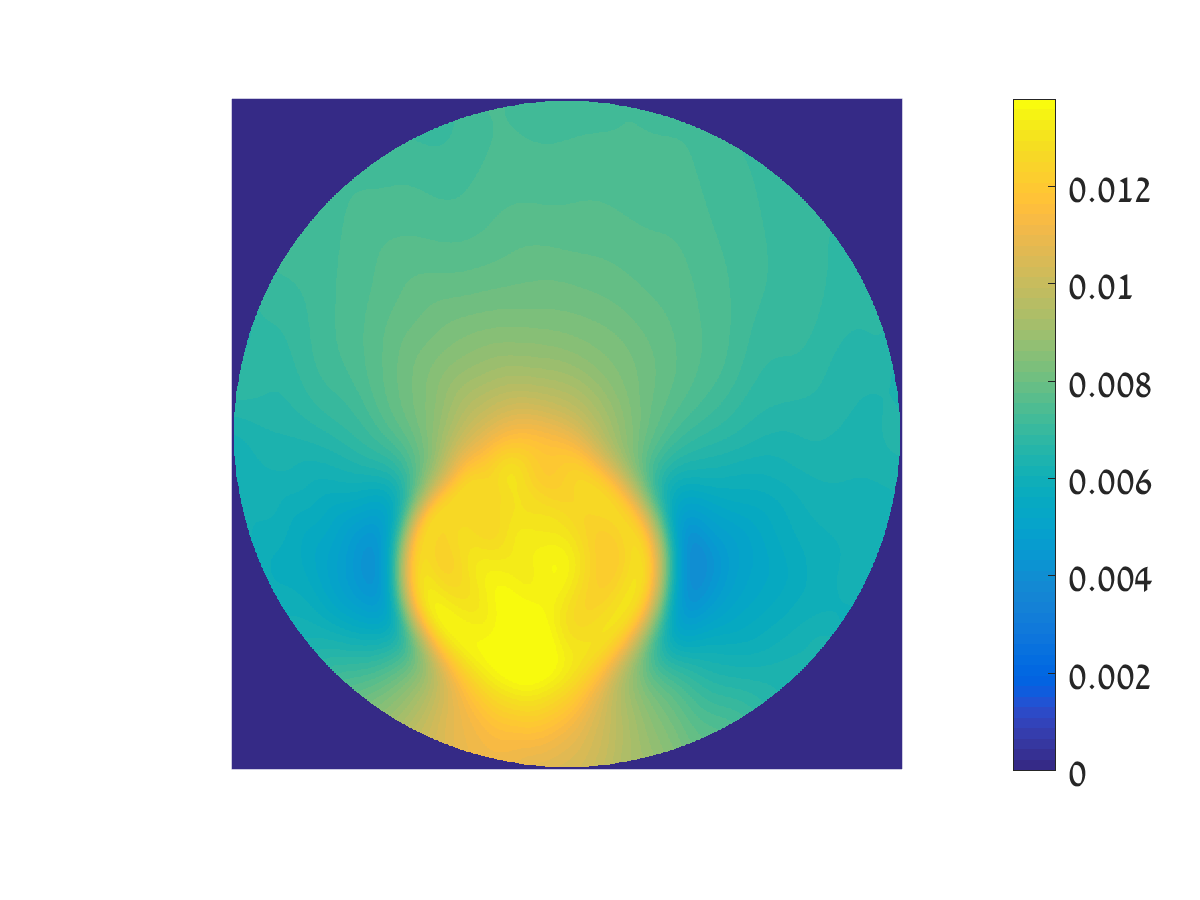}}
\subfigure[Proposed, $n=2$]{\includegraphics[width=.32\linewidth]{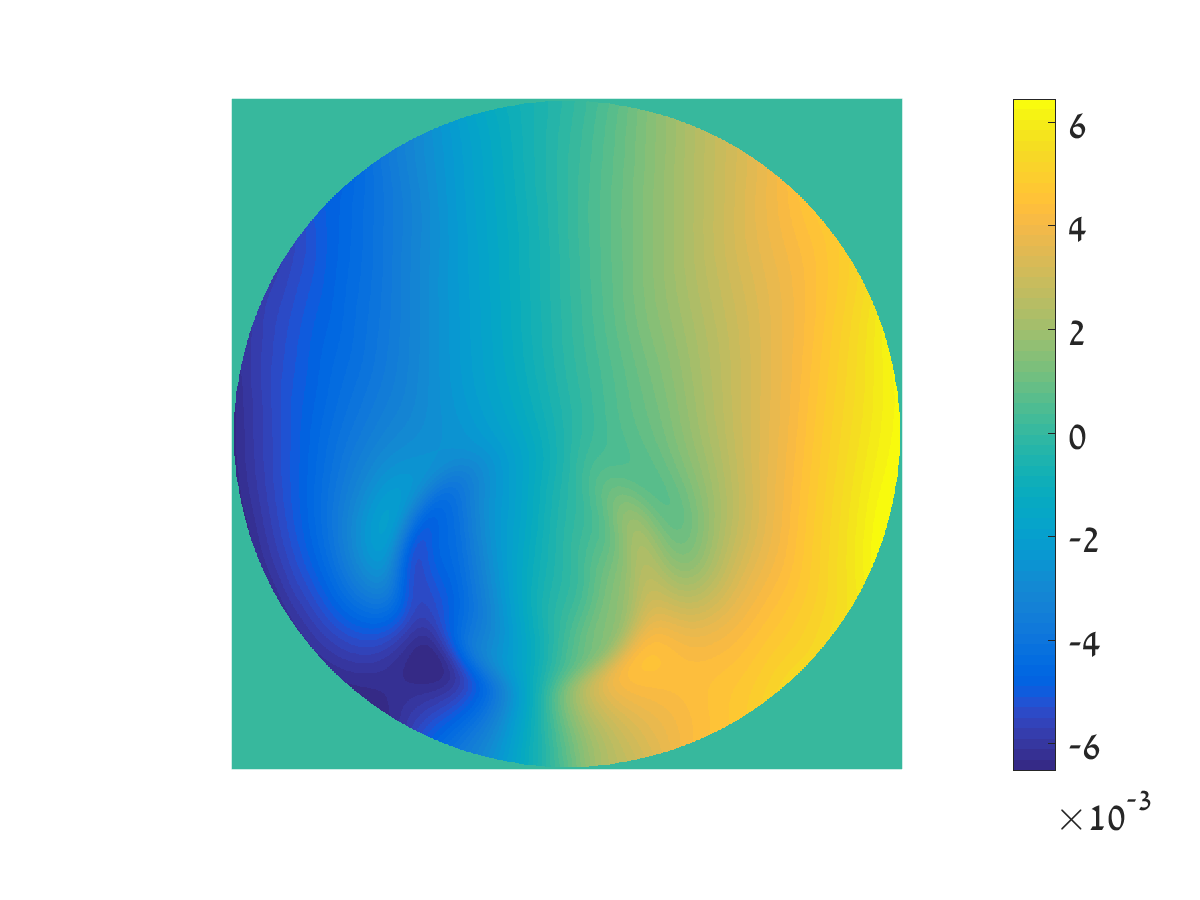}}
\subfigure[Proposed, $n=3$]{\includegraphics[width=.32\linewidth]{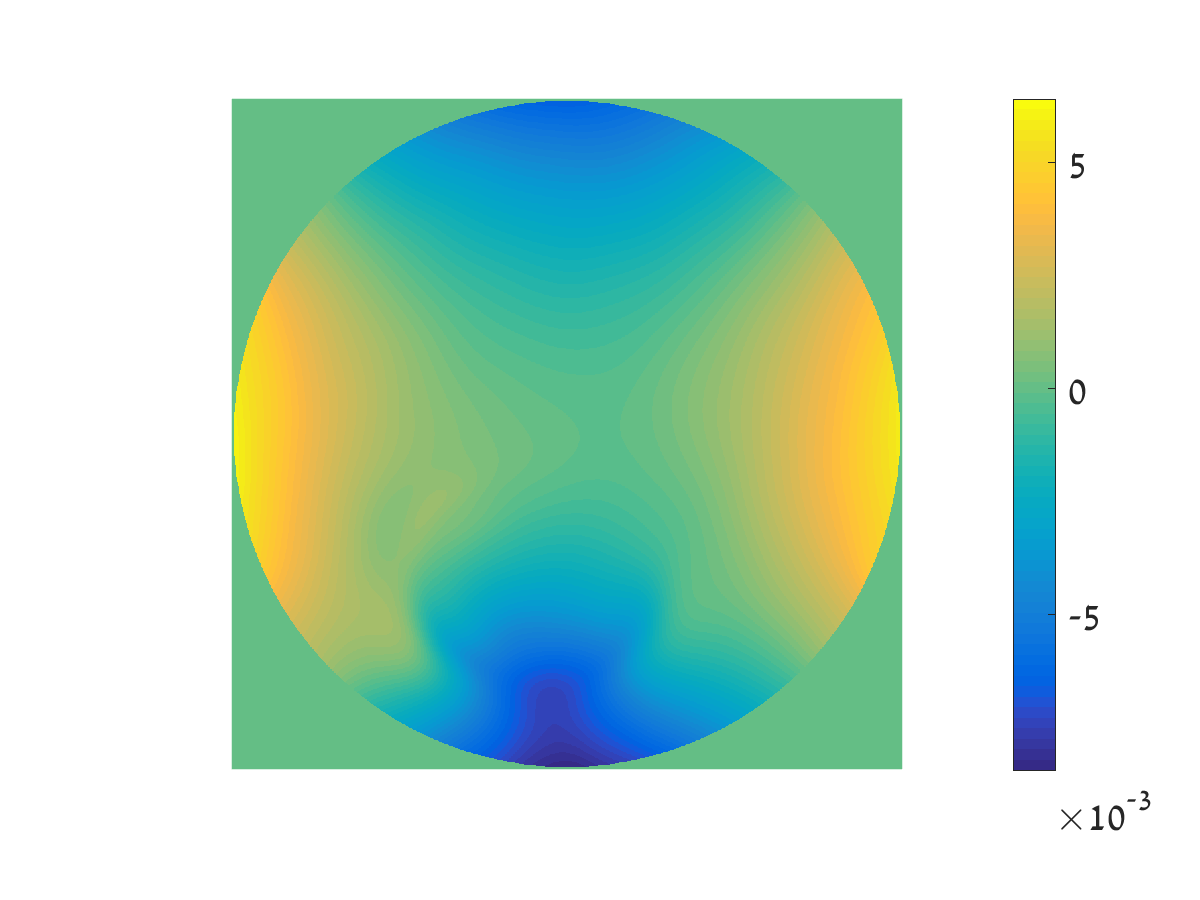}}\\[0.2cm]
\subfigure[Error, $n=1$]{\includegraphics[width=.32\linewidth]{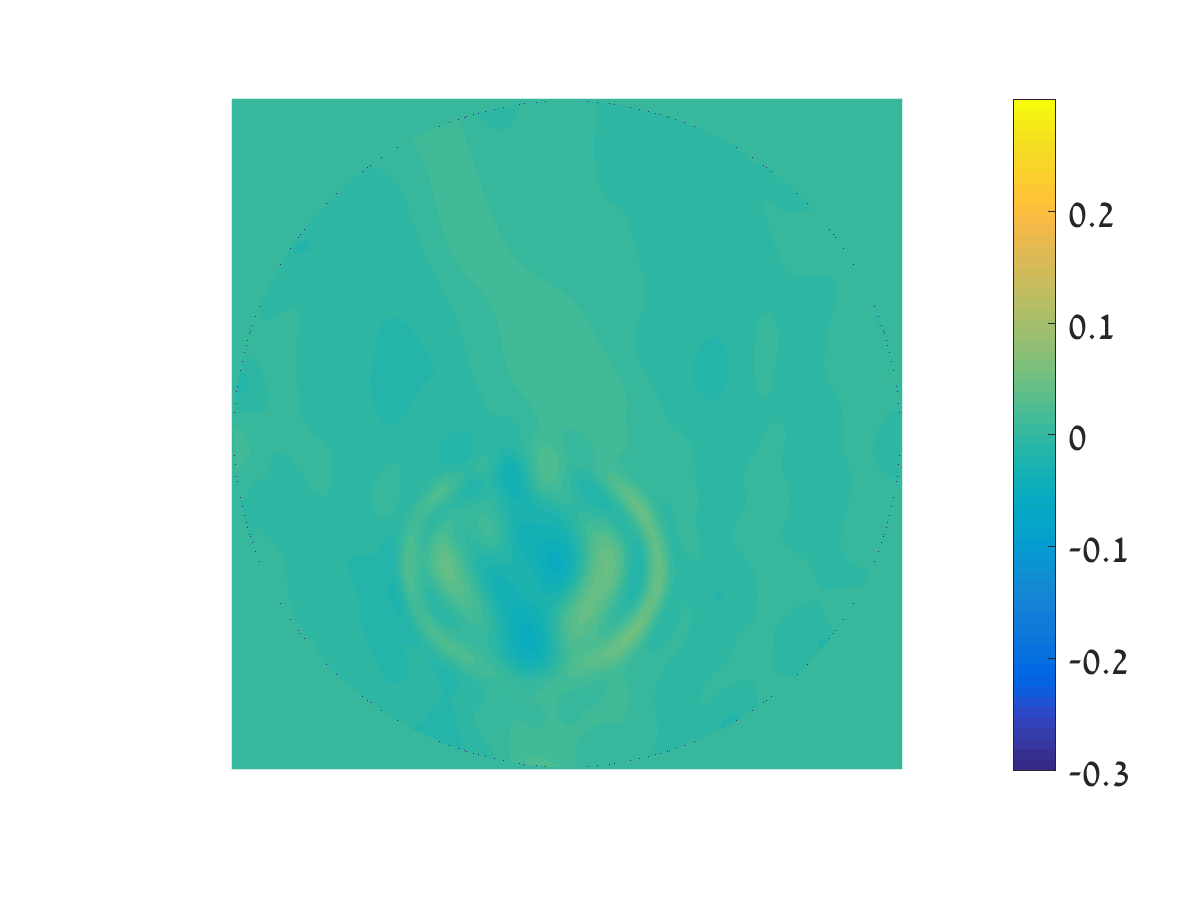}}
\subfigure[Error, $n=2$]{\includegraphics[width=.32\linewidth]{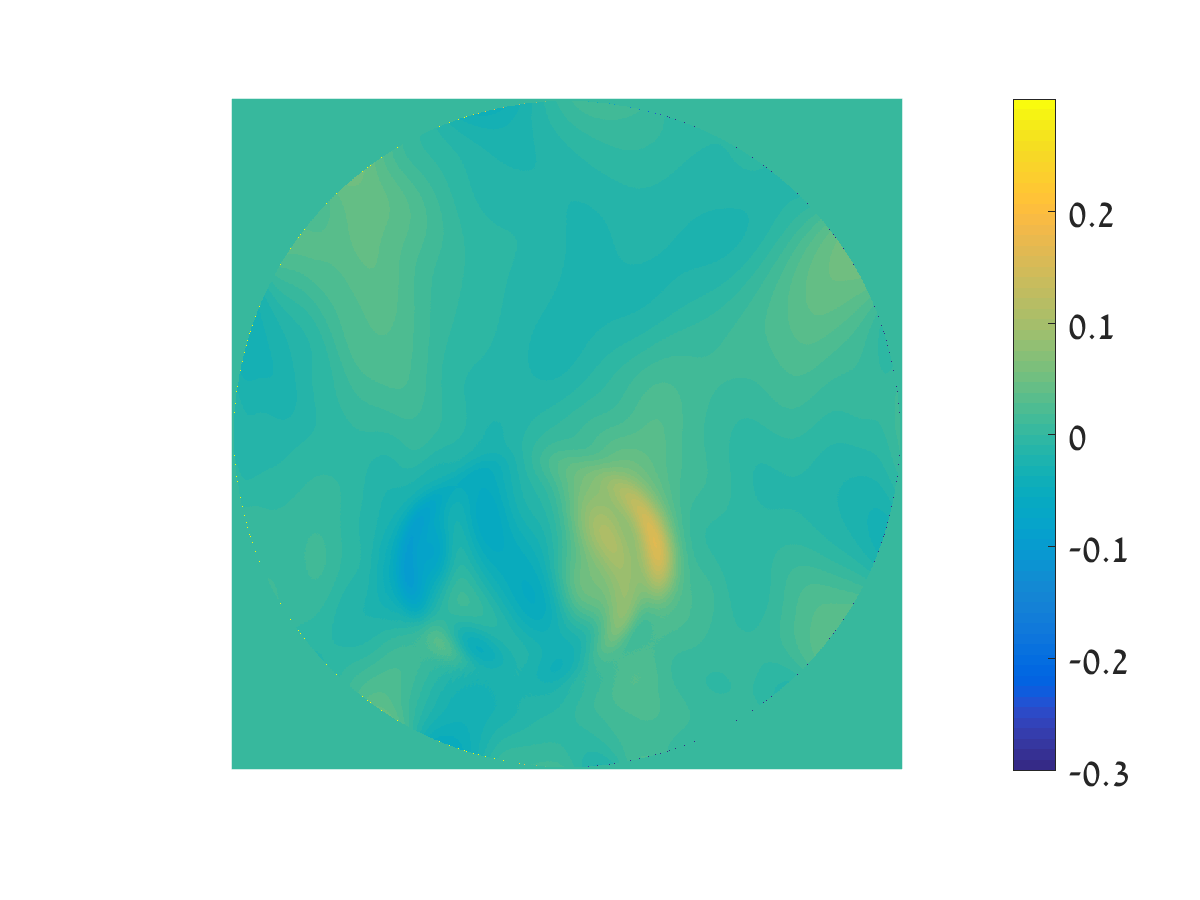}}
\subfigure[Error, $n=3$]{\includegraphics[width=.32\linewidth]{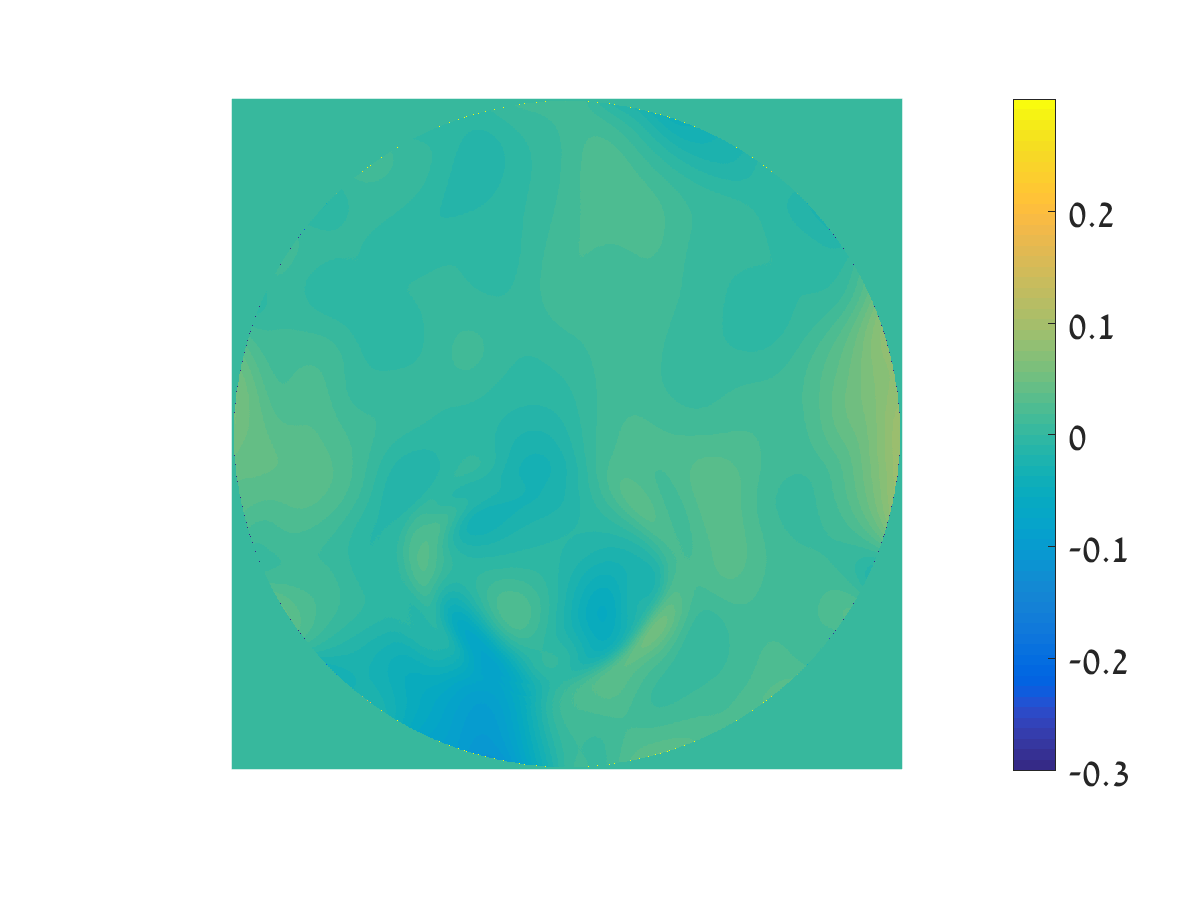}}
\end{center}
\caption{Top: ground truth (FEM) of $\partial u/\partial x$ for $n=1,2,3$ given phantom $1$. Middle: $\partial u/\partial x$ reconstruction by the proposed method. MSE =$(3.77e-8,3.20e-8,2.84e-8)$ PSNR = $(37.03,31.22,34.02)$. Bottom: relative error.}
\label{fig:ux1_res}
\end{figure}

Figure~\ref{fig:ux1_res} shows the derivative of $u$ with respect to $x$. The top row is the finite difference approximation of the FEM result. The middle row is the analytical derivative of our result, and the bottom row shows the relative error.

\begin{figure}
\begin{center}
\includegraphics[width=.4\linewidth]{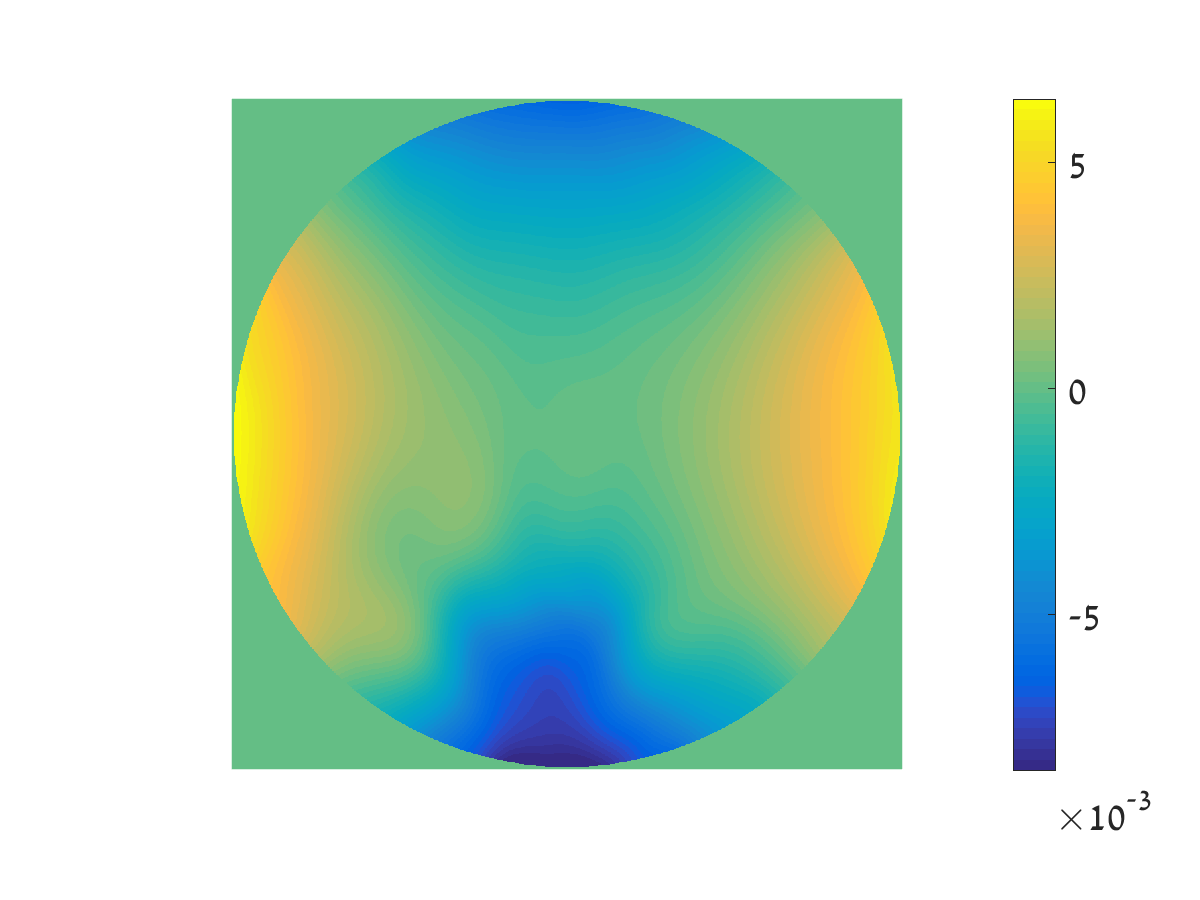}
\includegraphics[width=.4\linewidth]{UxL1H02n3.png}
\end{center}
\caption{The derivative $\partial u/\partial x$ of phantom $1$ for $n=3$. Left: without $L_\infty$ norm. MSE = 3.93e-8, PSNR = 32.61. Right: with $L_\infty$ norm. MSE = 2.84e-8, PSNR = 34.02}
\label{fig:compare}
\end{figure}

We further demonstrate the effect of the $L_\infty$ norm in the cost function. The left and right images of Figure~\ref{fig:compare} stand for the derivative $\partial u/\partial x$ of phantom $1$ without and with the $L_\infty$ term respectively. Clearly, this additional norm yields better reconstruction both visually (sharper circle edges) and quantitatively. 

\begin{figure}
\begin{center}
\includegraphics[width=.6\linewidth]{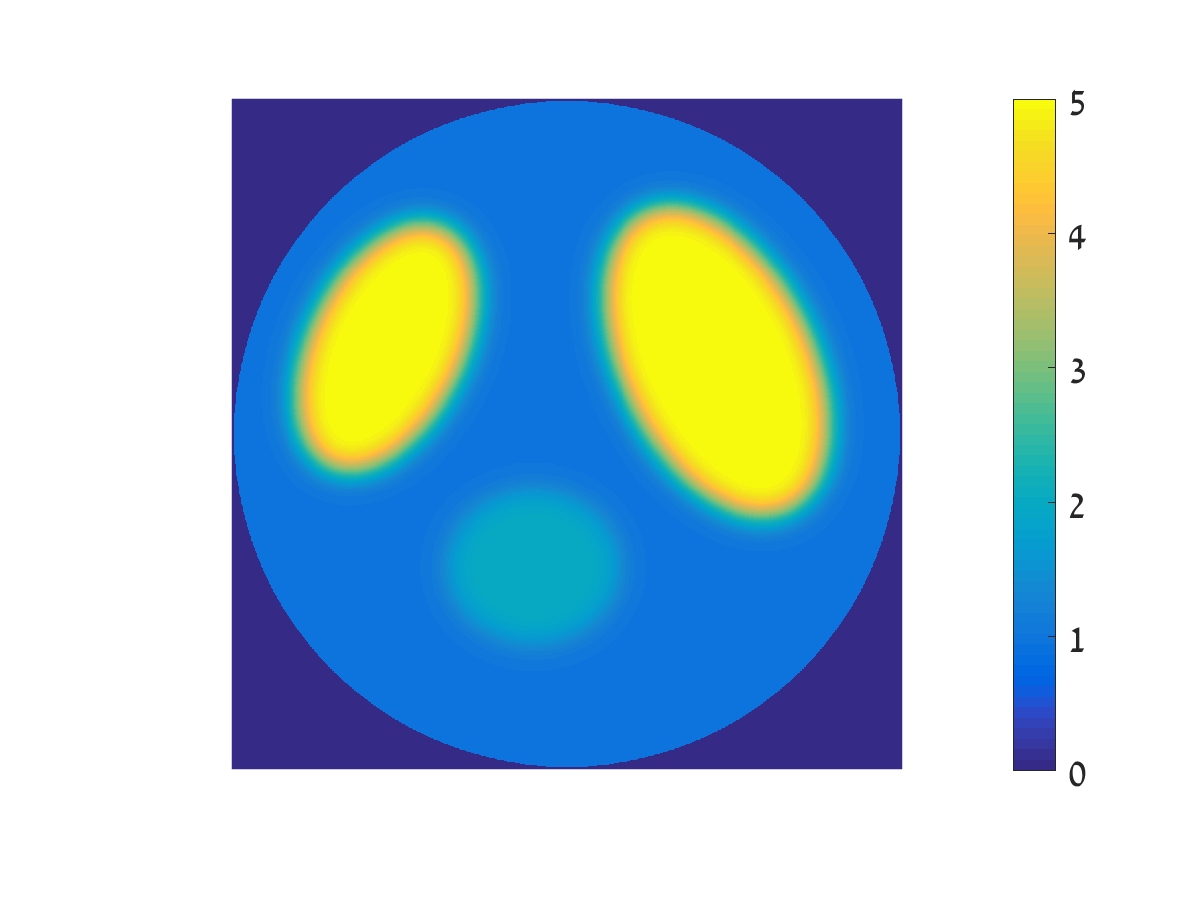}
\end{center}
\caption{The conductivity $\sigma$ of phantom $2$}
\label{fig:sig2}
\end{figure}

\begin{figure}
\begin{center}
\subfigure[FEM, $n=1$]{\includegraphics[width=.32\linewidth]{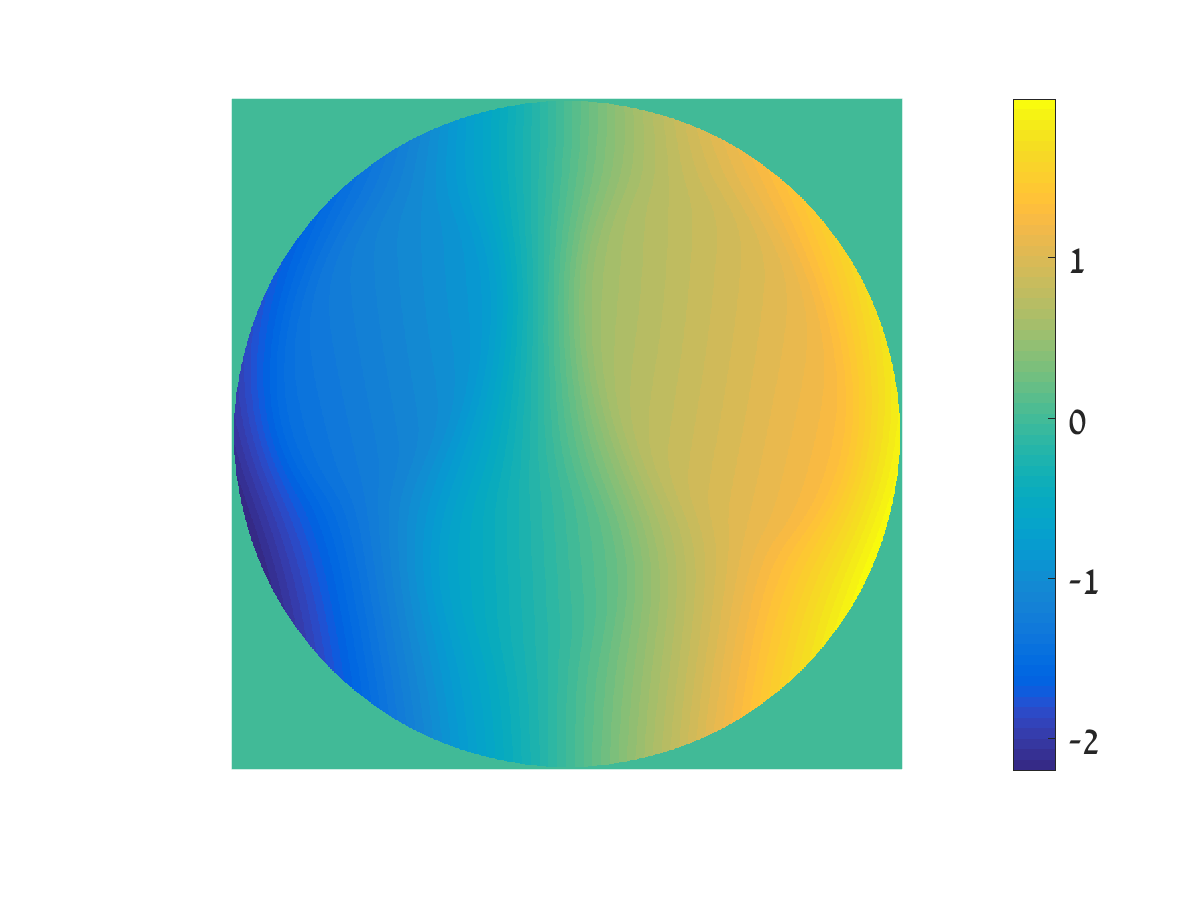}}
\subfigure[FEM, $n=2$]{\includegraphics[width=.32\linewidth]{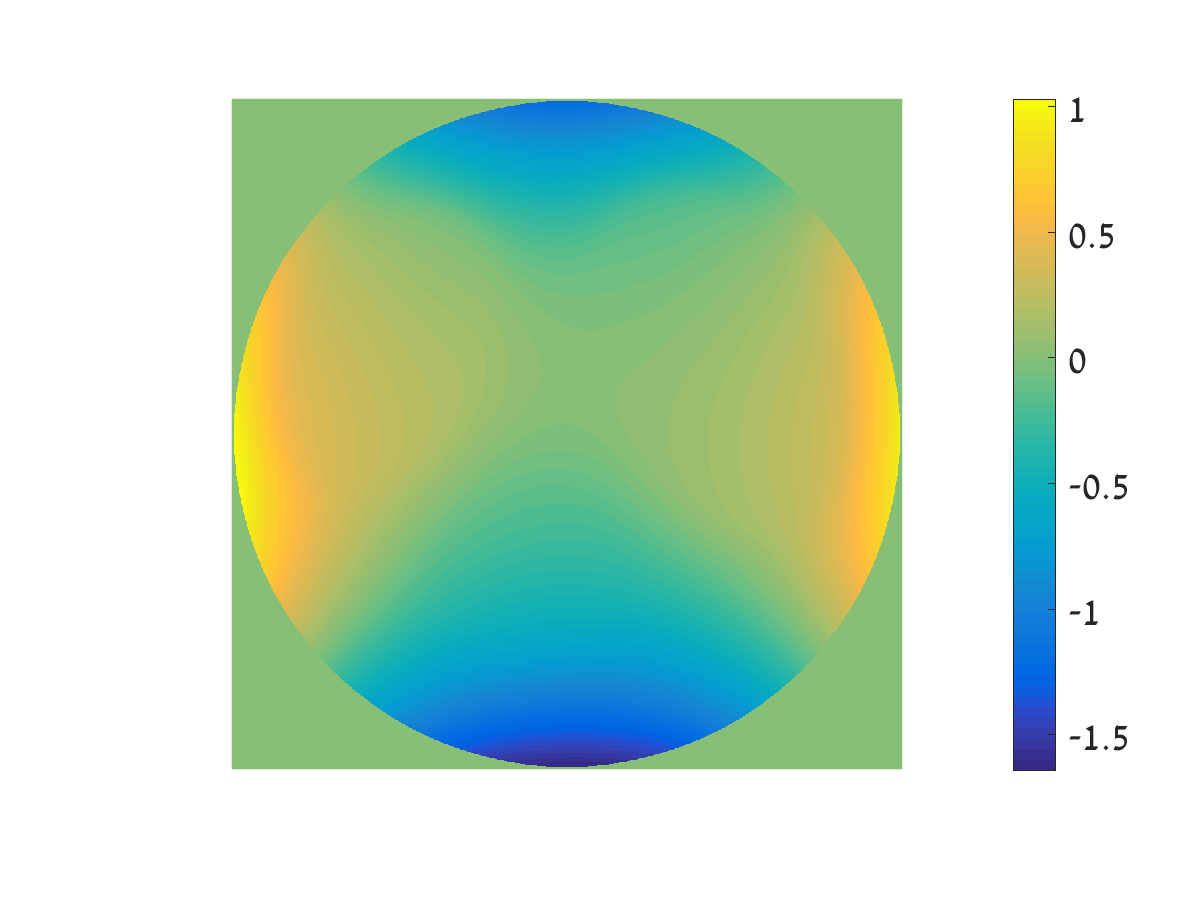}}
\subfigure[FEM, $n=3$]{\includegraphics[width=.32\linewidth]{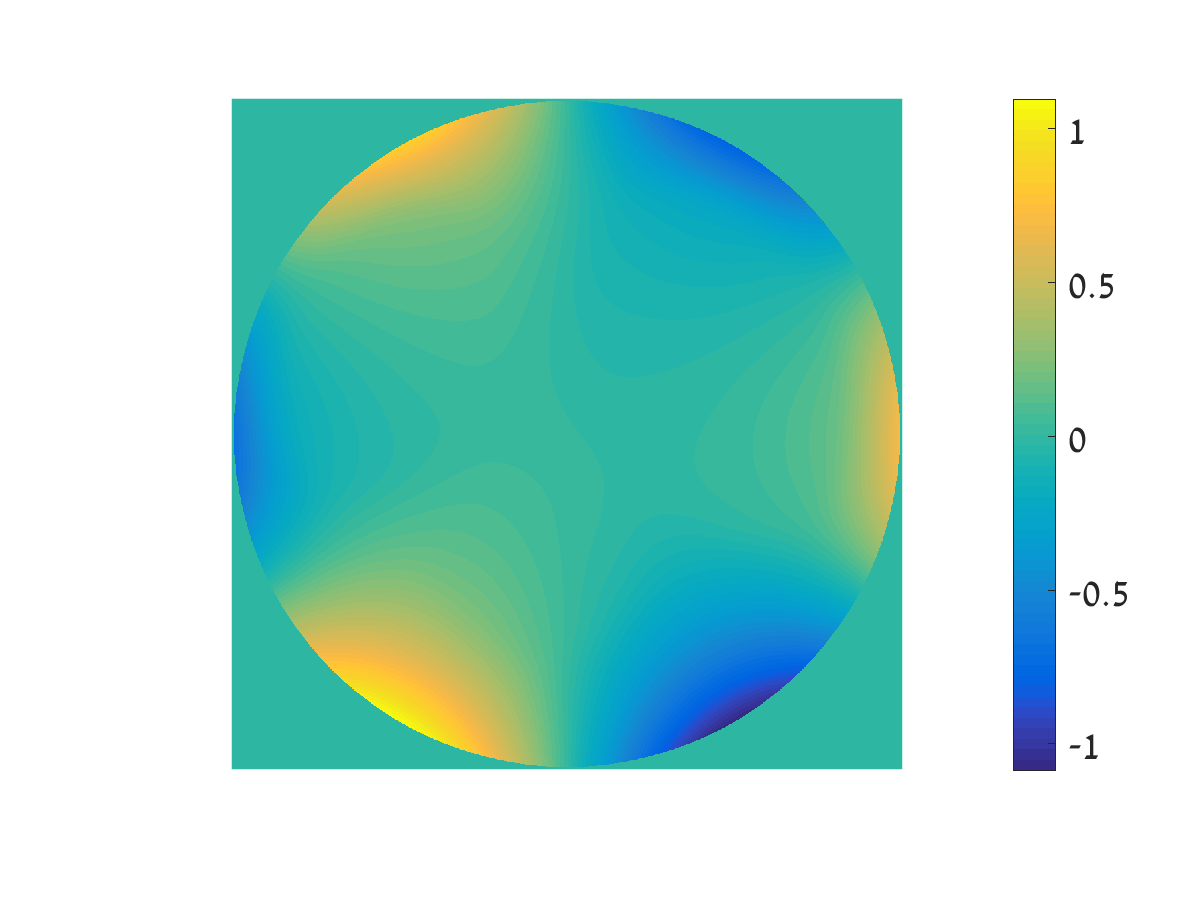}}\\[0.2cm]
\subfigure[Proposed, $n=1$]{\includegraphics[width=.32\linewidth]{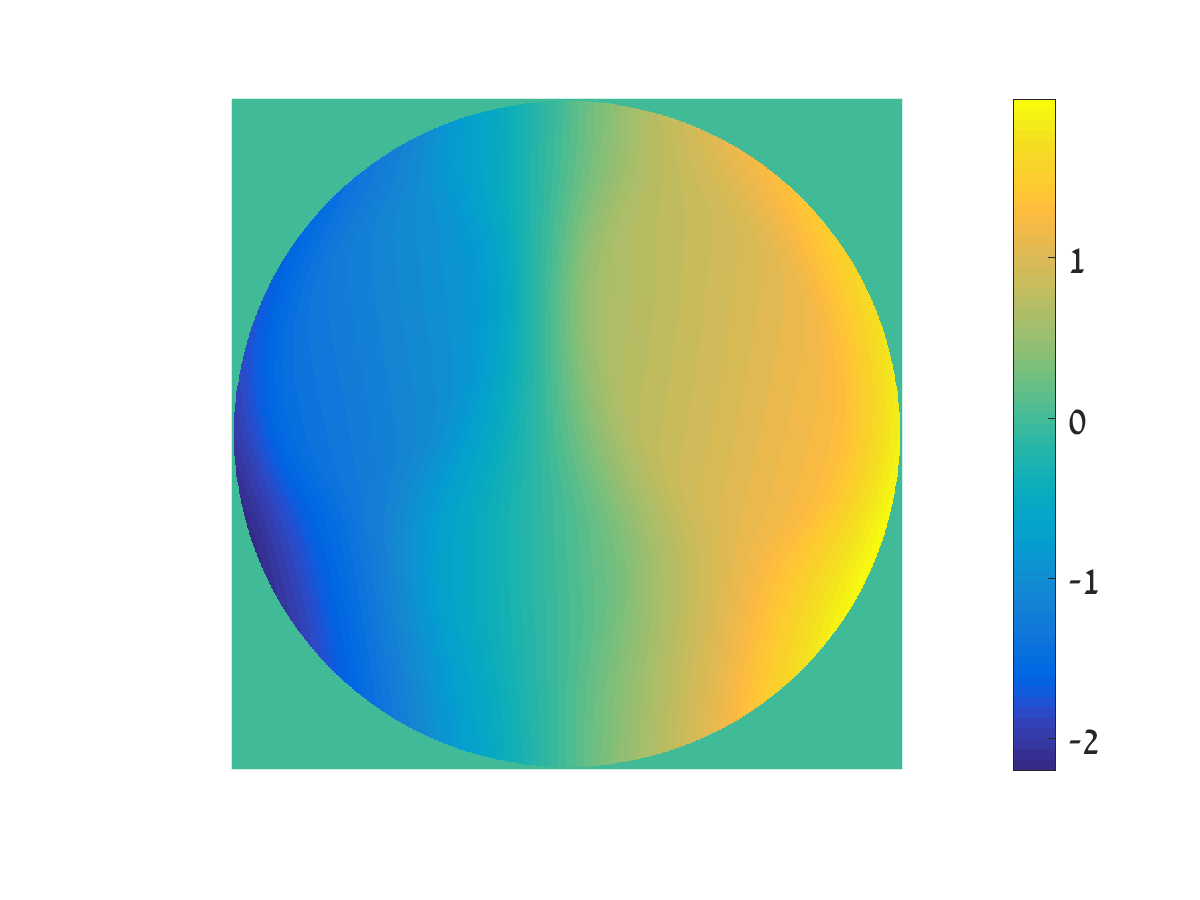}}
\subfigure[Proposed, $n=2$]{\includegraphics[width=.32\linewidth]{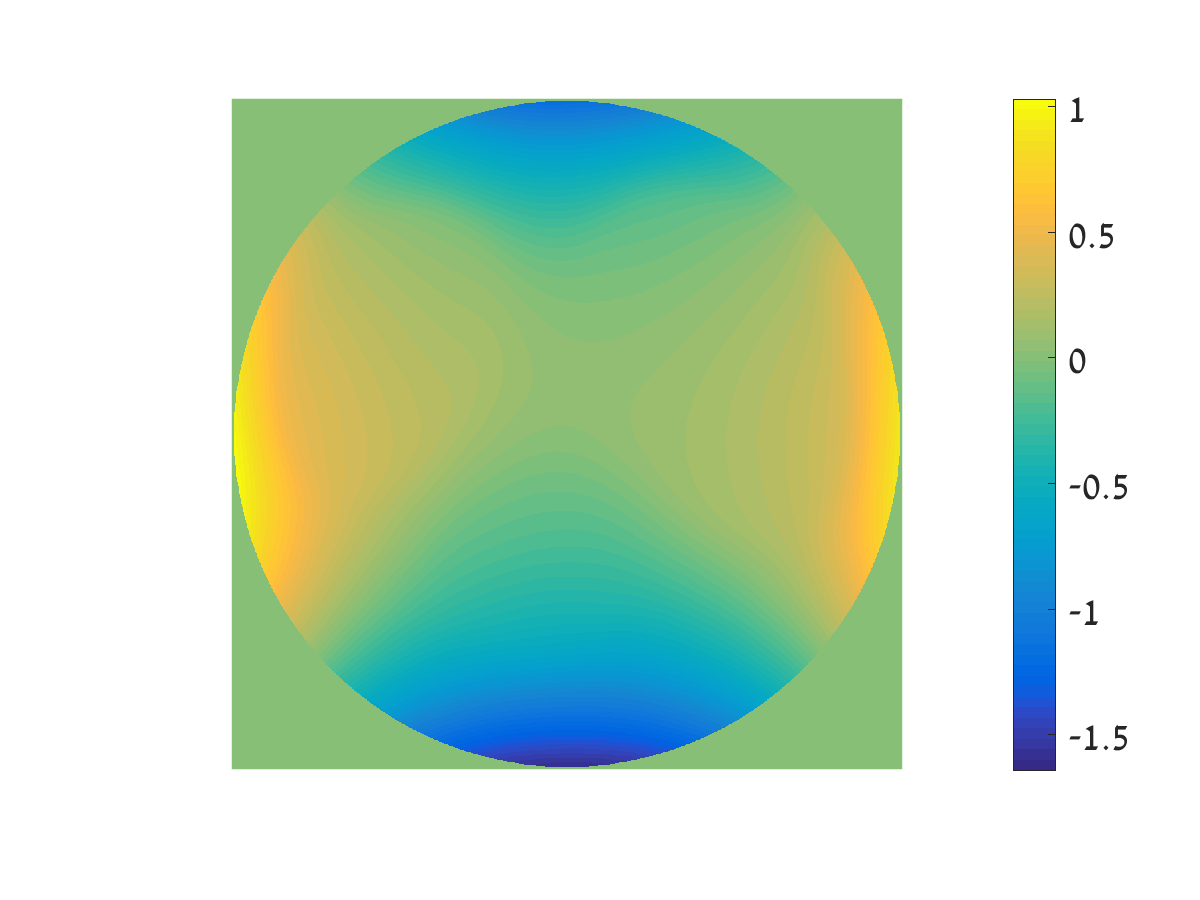}}
\subfigure[Proposed, $n=3$]{\includegraphics[width=.32\linewidth]{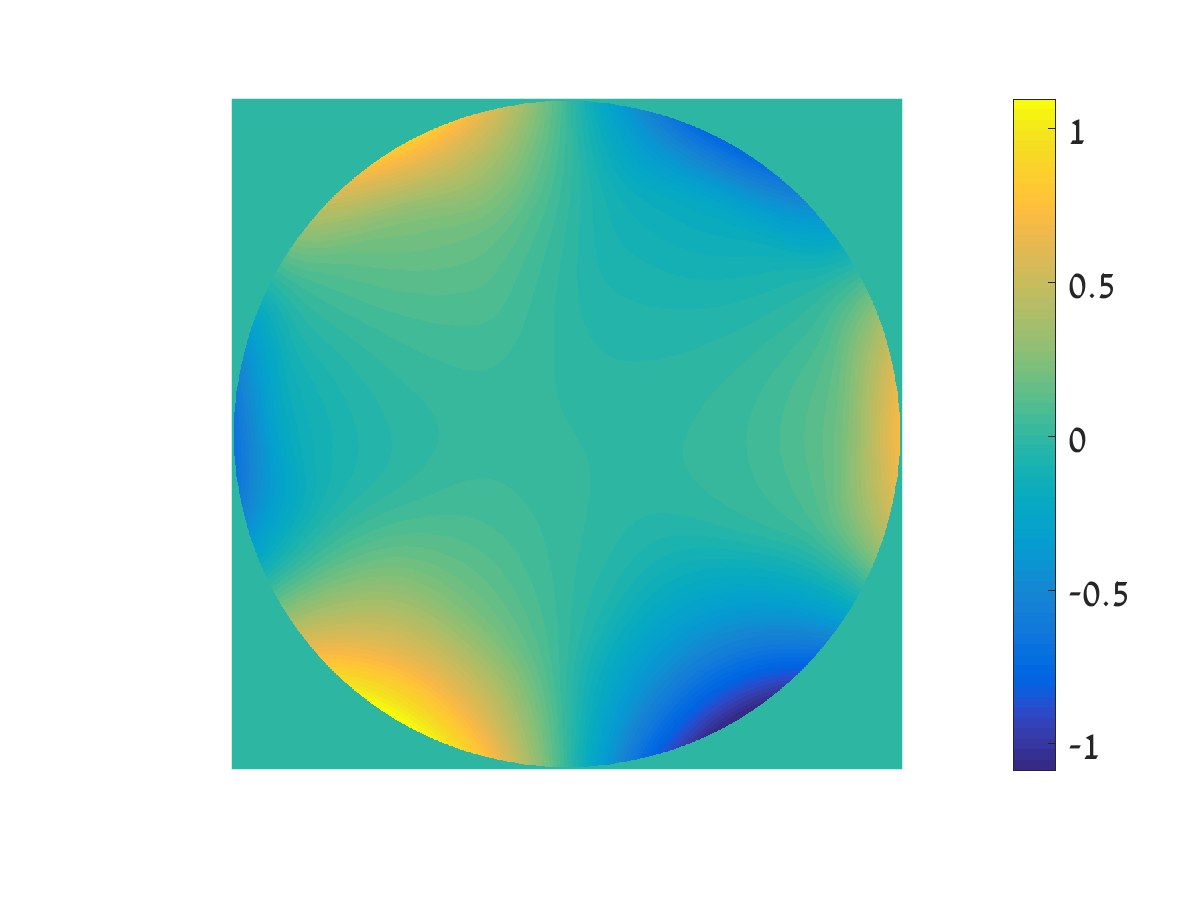}}\\[0.2cm]
\subfigure[Error, $n=1$]{\includegraphics[width=.32\linewidth]{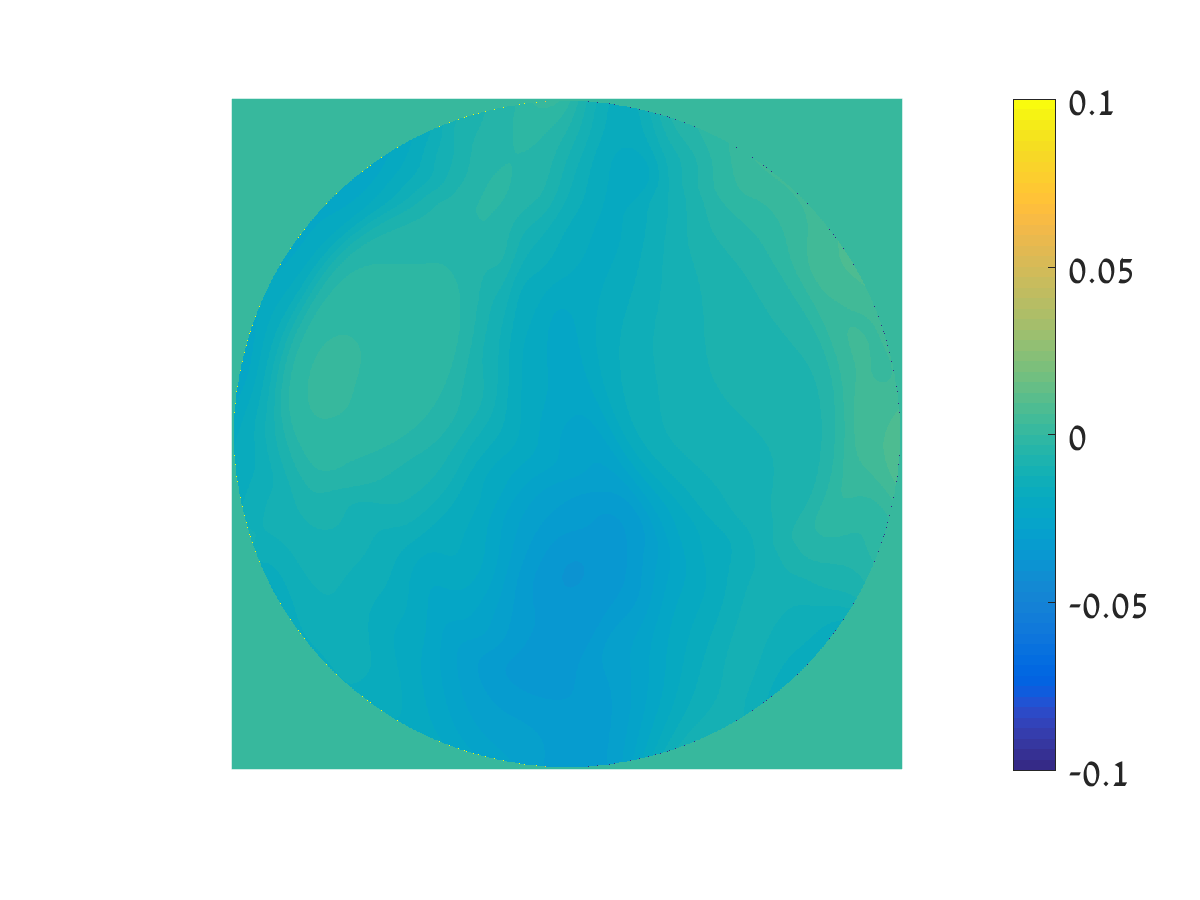}}
\subfigure[Error, $n=2$]{\includegraphics[width=.32\linewidth]{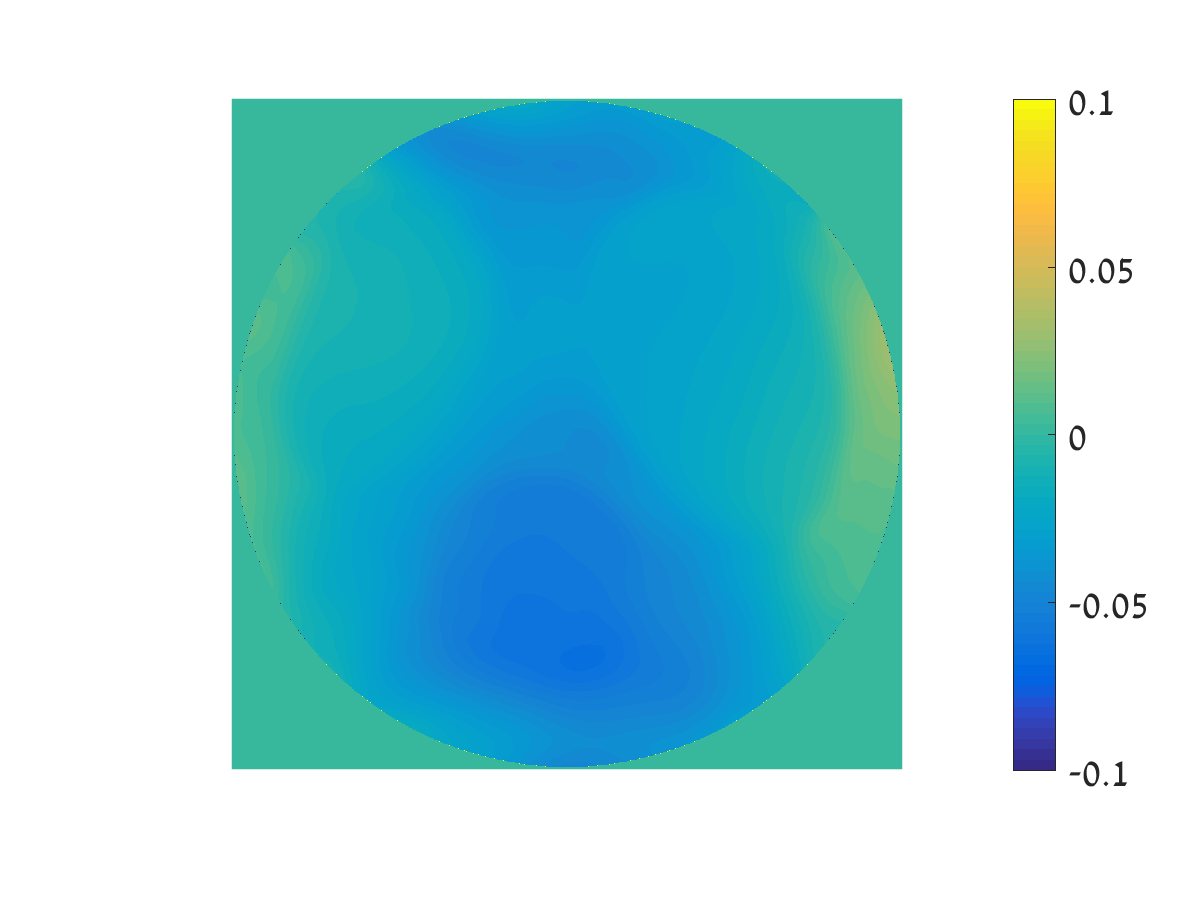}}
\subfigure[Error, $n=3$]{\includegraphics[width=.32\linewidth]{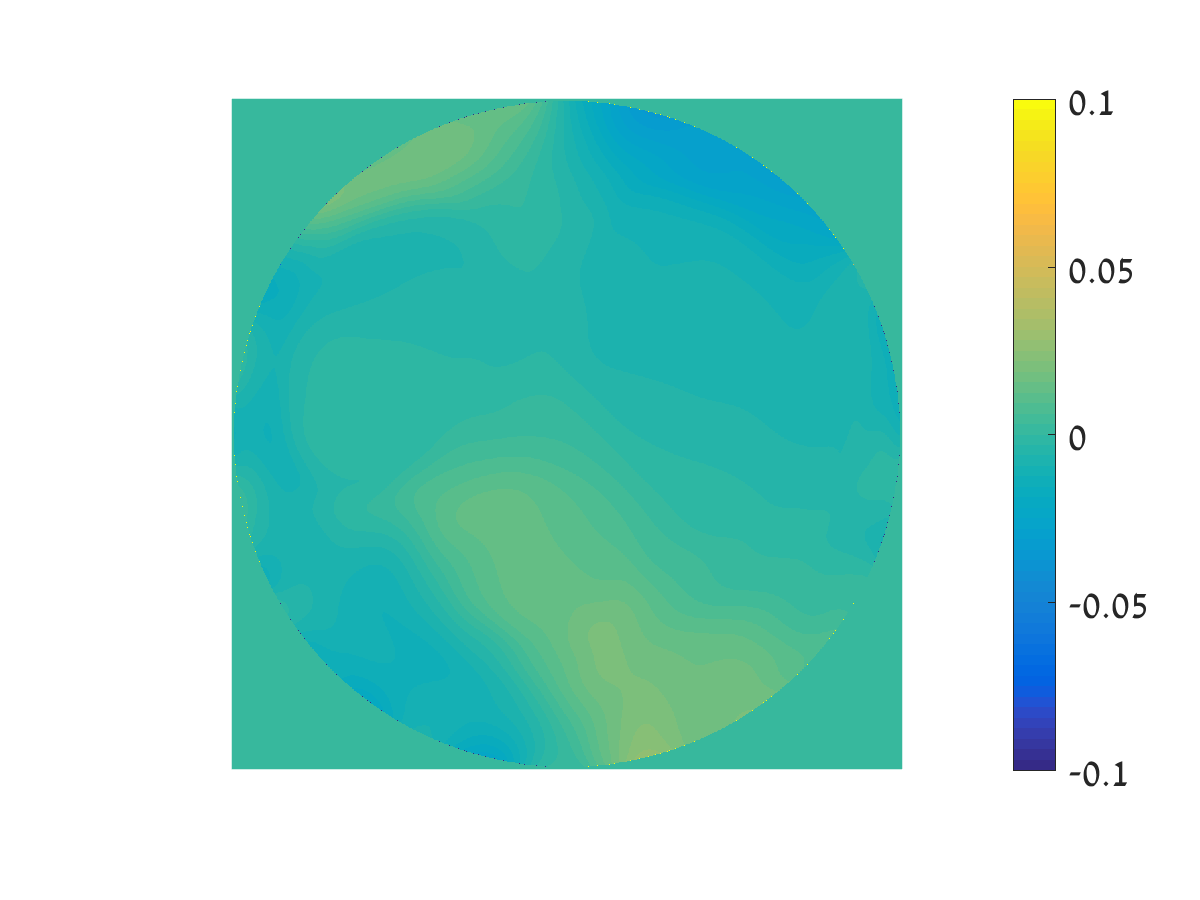}}
\end{center}
\caption{Top: ground truth (FEM) of $u$ for $n=1,2,3$ given phantom $2$. Middle: reconstruction by the proposed method. MSE = $(1.72e-3,1.22e-3,2.35e-4)$ PSNR=$(34.49,33.46,37.06)$. Bottom: relative error.}
\label{fig:u2_res}
\end{figure}

\begin{figure}
\begin{center}
\subfigure[FEM, $n=1$]{\includegraphics[width=.32\linewidth]{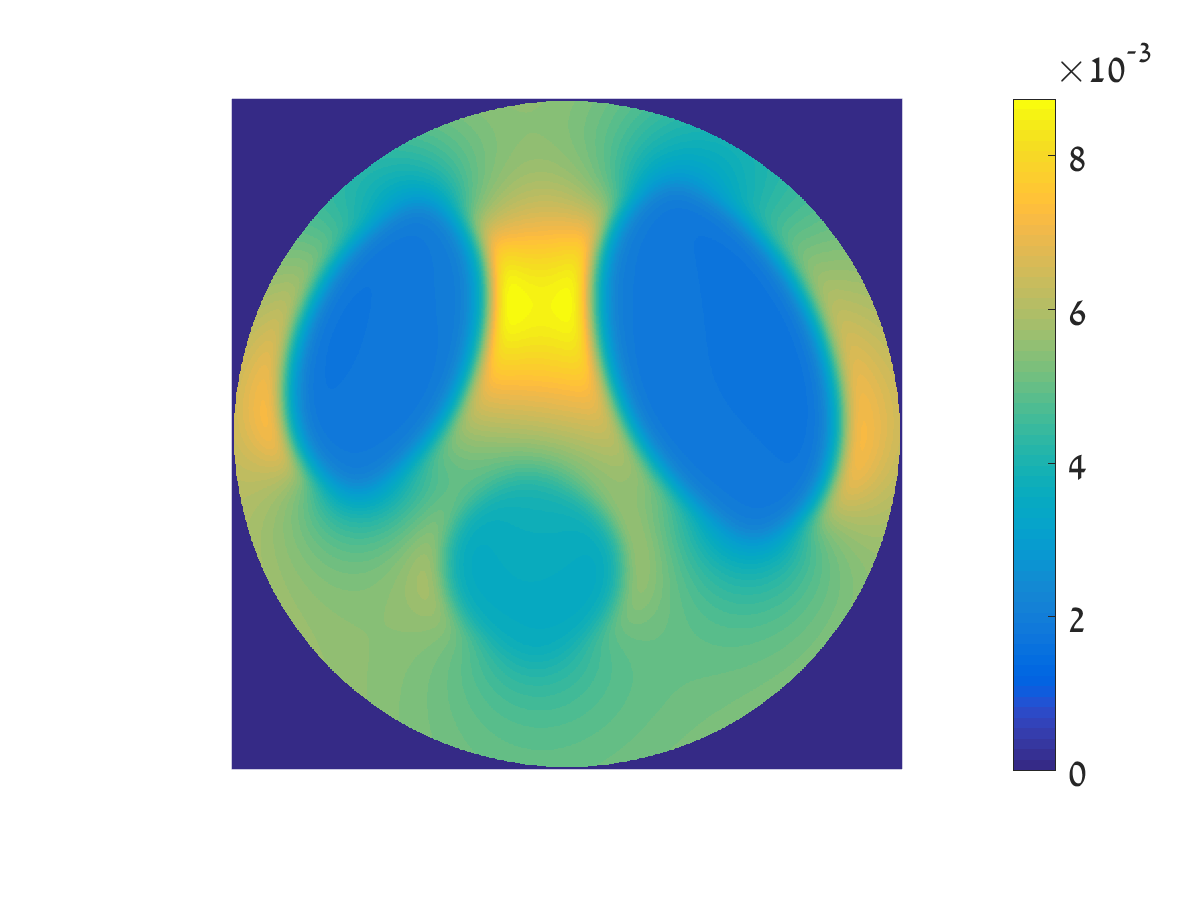}}
\subfigure[FEM, $n=2$]{\includegraphics[width=.32\linewidth]{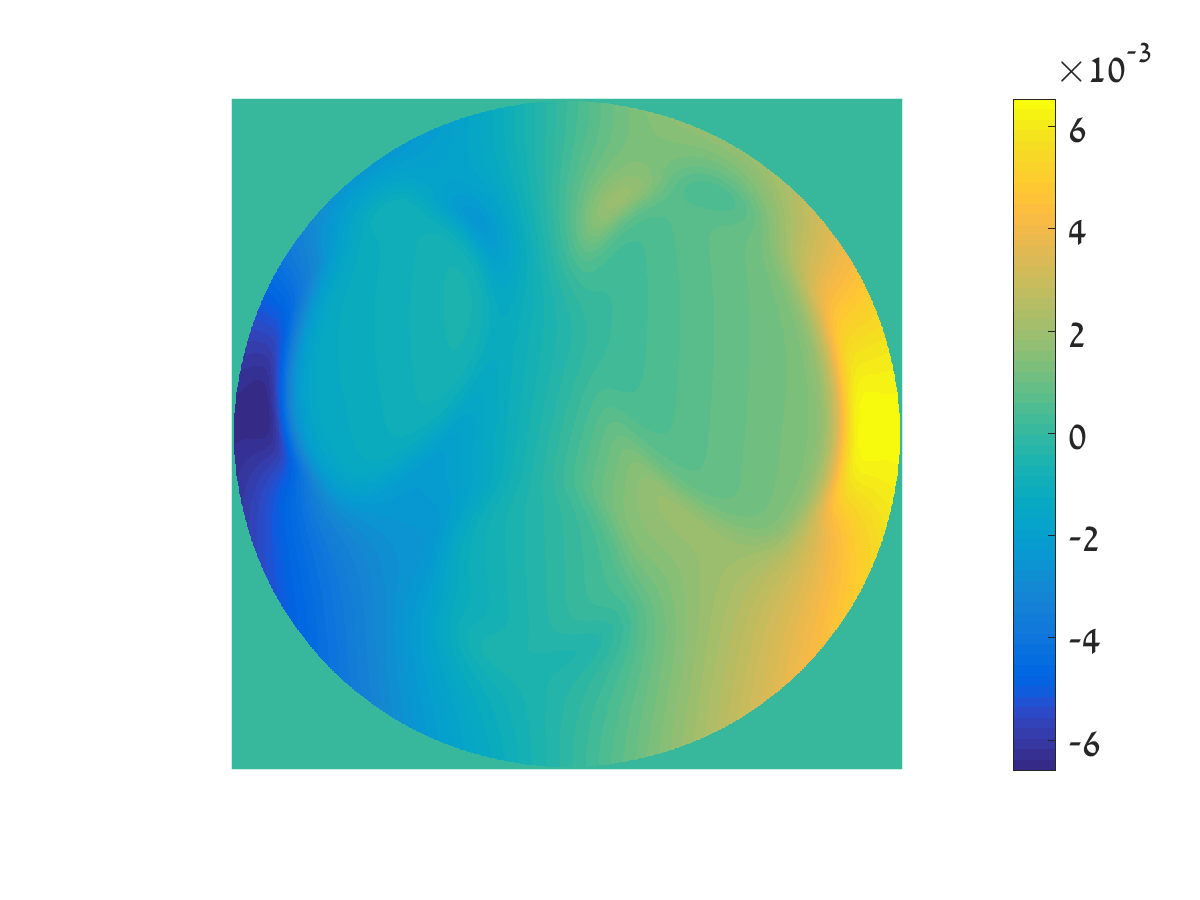}}
\subfigure[FEM, $n=3$]{\includegraphics[width=.32\linewidth]{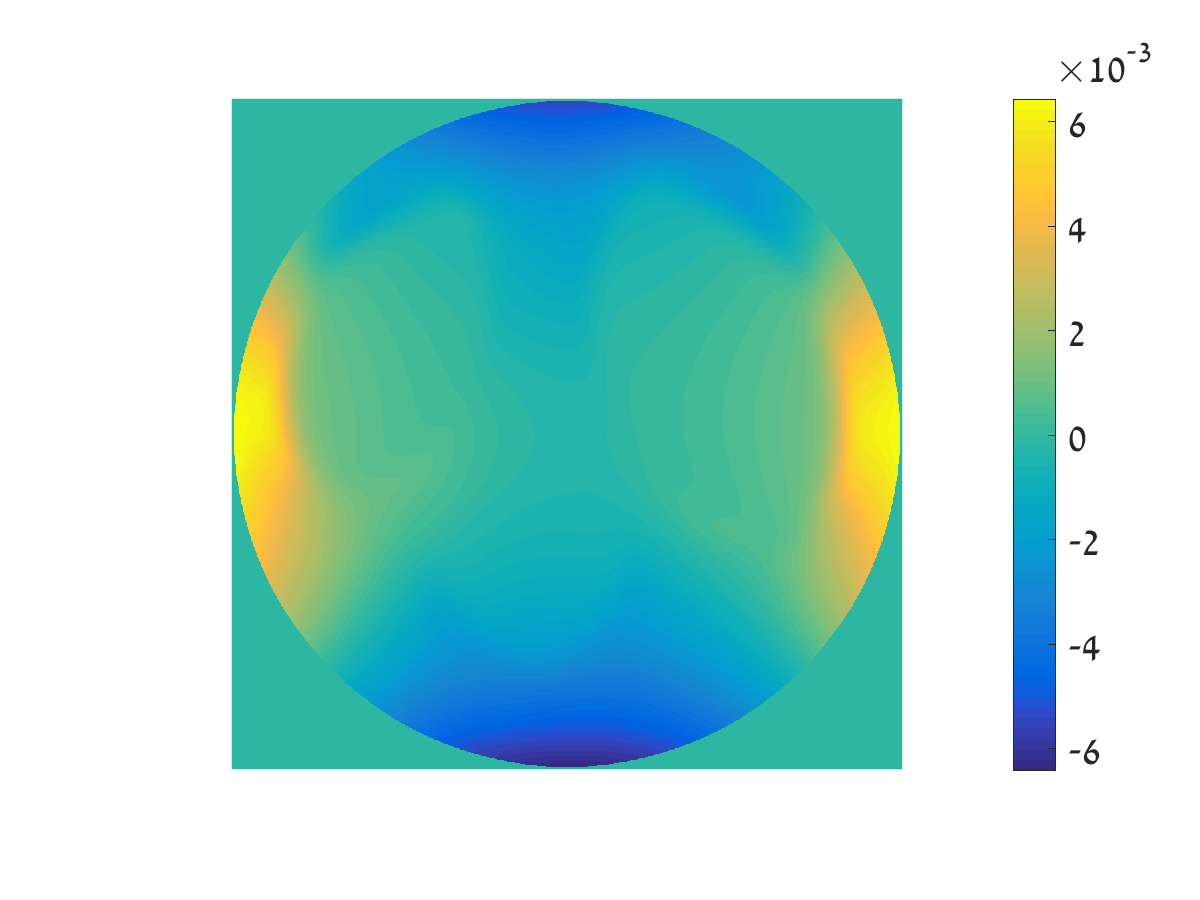}}\\[0.2cm]
\subfigure[Proposed, $n=1$]{\includegraphics[width=.32\linewidth]{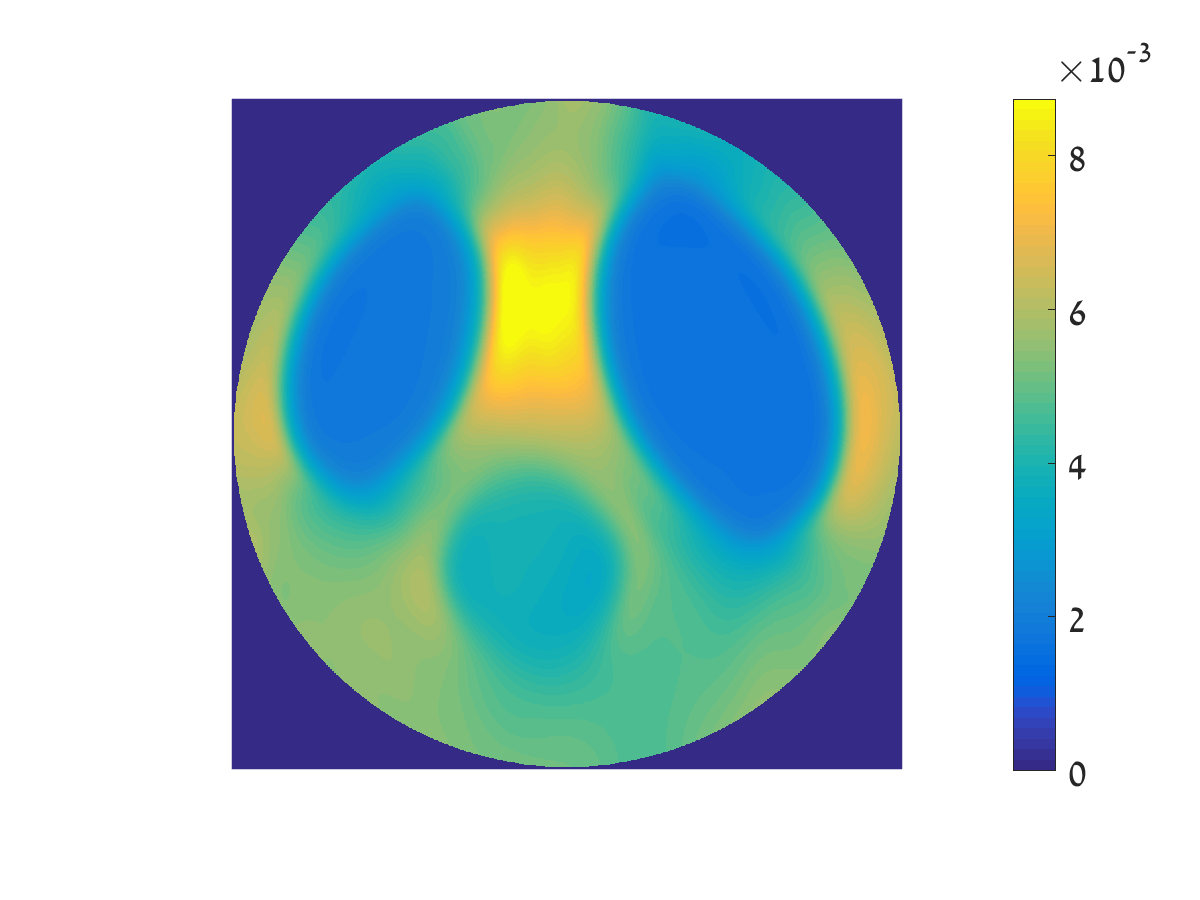}}
\subfigure[Proposed, $n=2$]{\includegraphics[width=.32\linewidth]{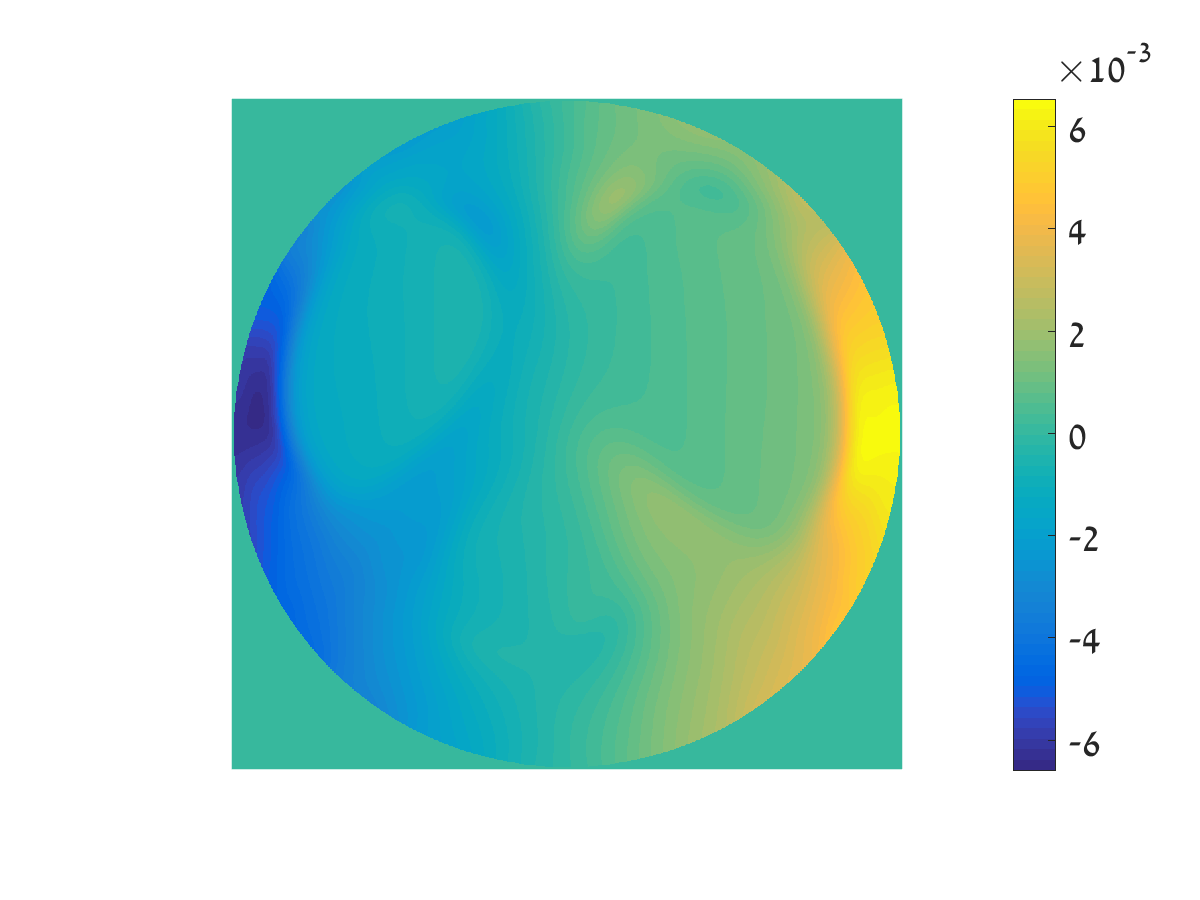}}
\subfigure[Proposed, $n=3$]{\includegraphics[width=.32\linewidth]{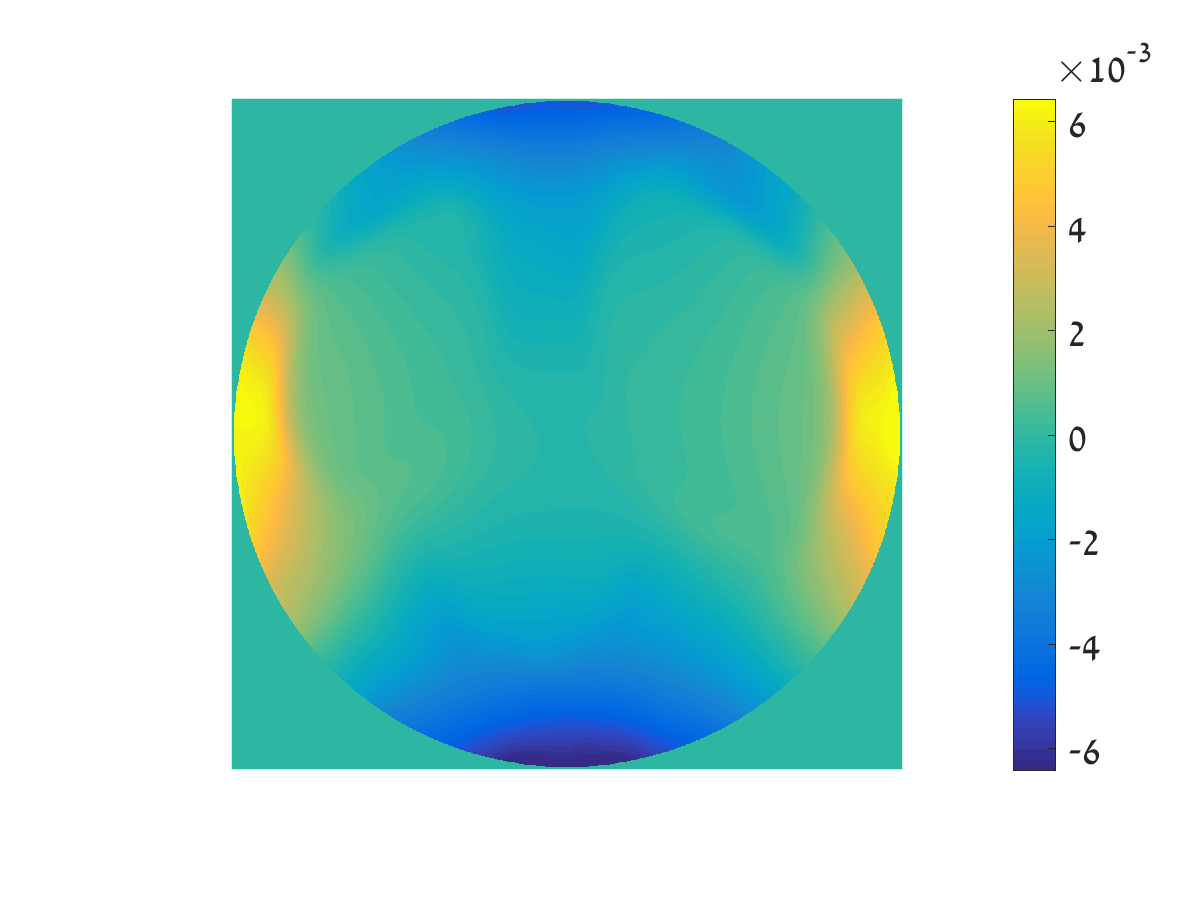}}\\[0.2cm]
\subfigure[Error, $n=1$]{\includegraphics[width=.32\linewidth]{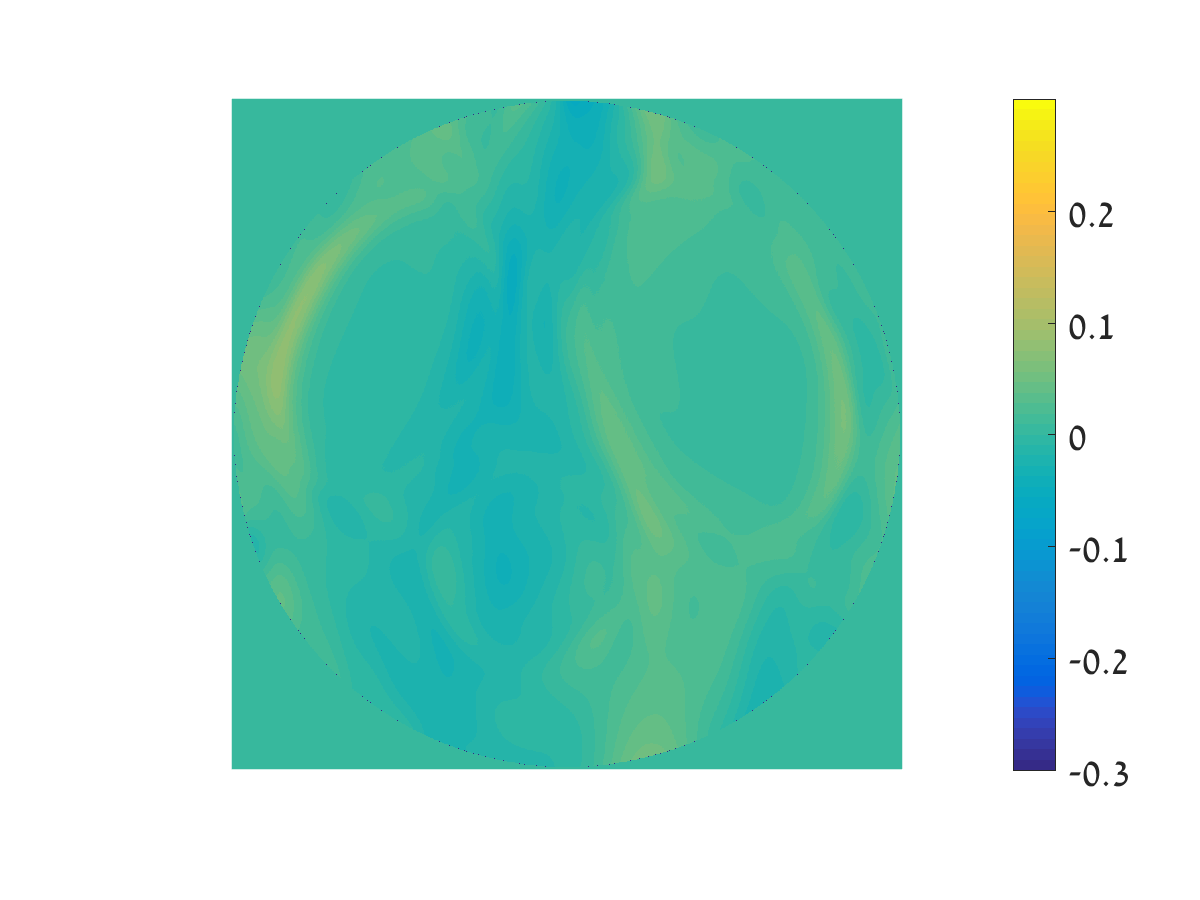}}
\subfigure[Error, $n=2$]{\includegraphics[width=.32\linewidth]{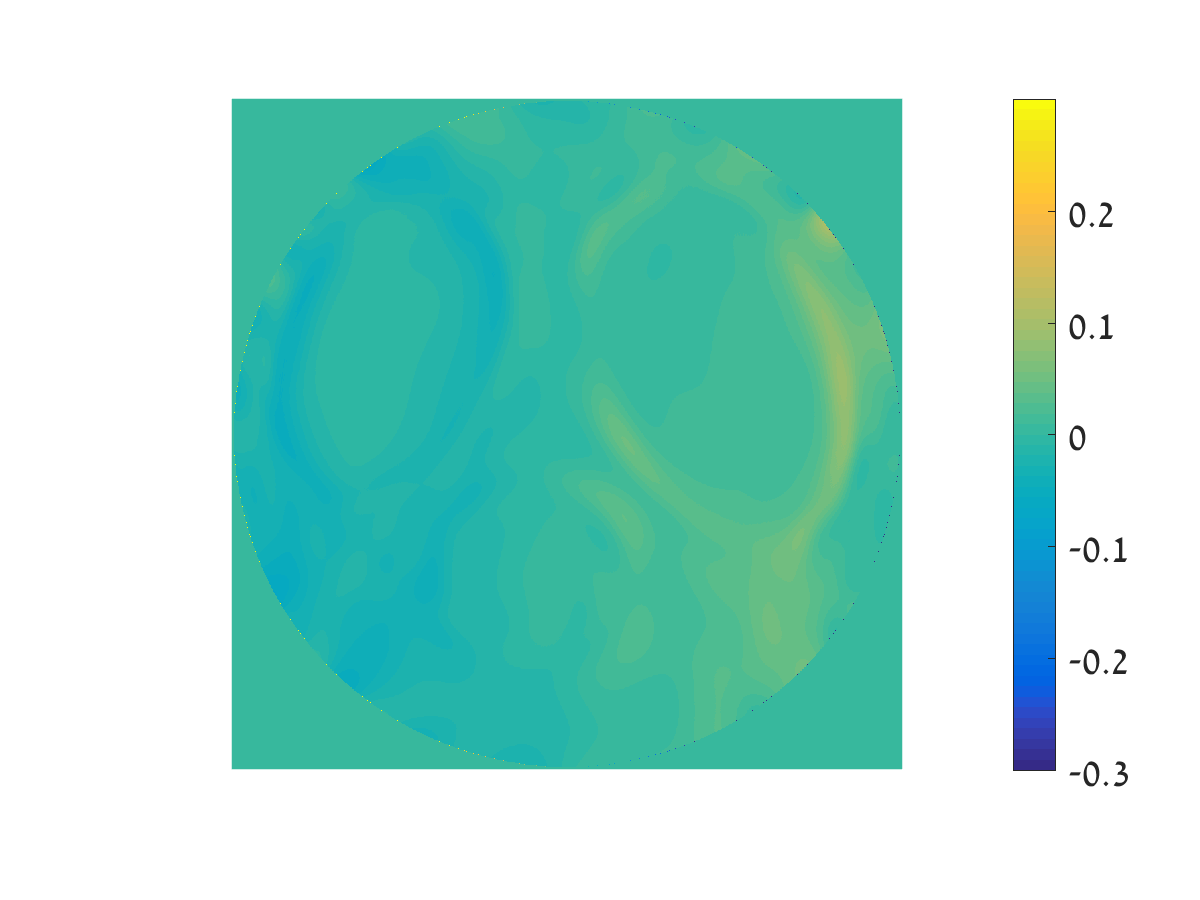}}
\subfigure[Error, $n=3$]{\includegraphics[width=.32\linewidth]{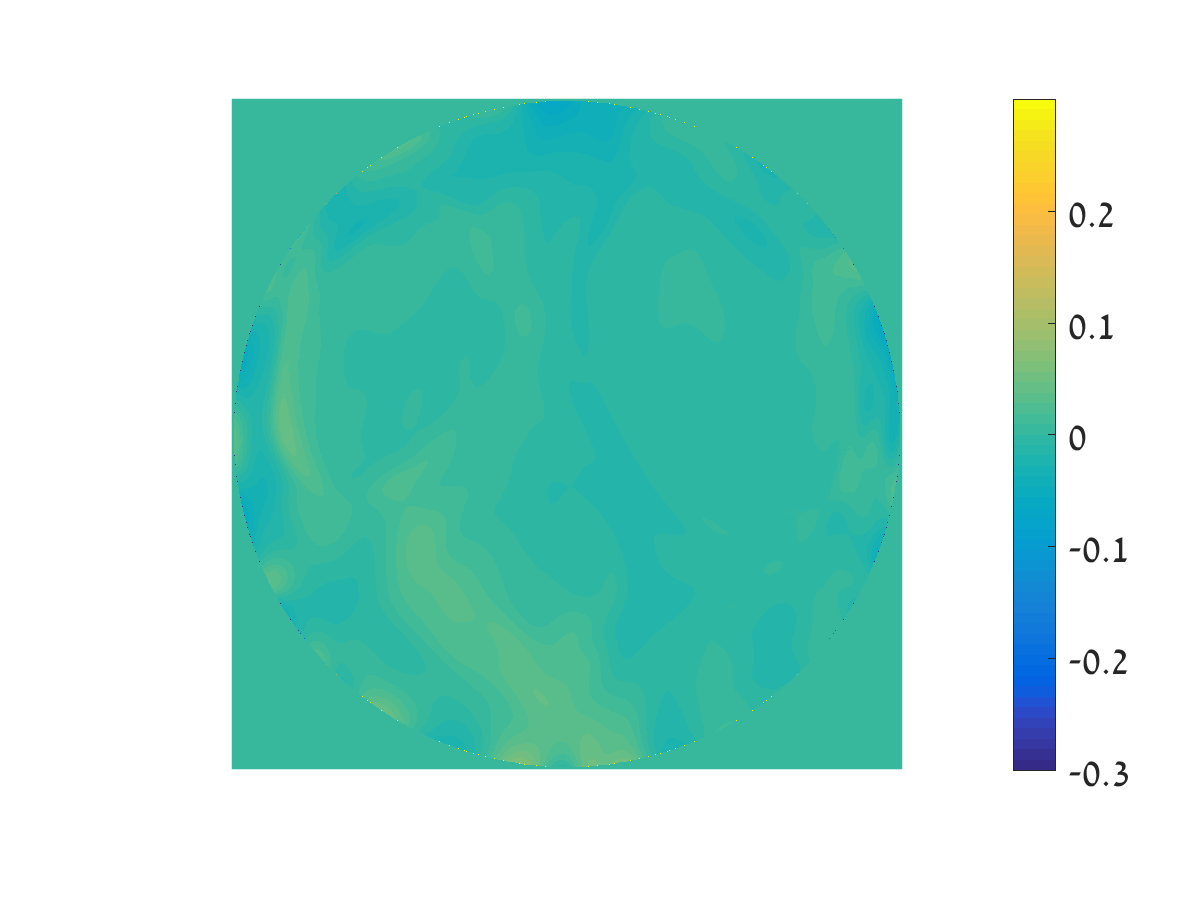}}
\end{center}
\caption{Top: ground truth (FEM) of $\partial u/\partial x$ for $n=1,2,3$ given phantom $2$. Middle: $\partial u/\partial x$ reconstruction by the proposed method. MSE=$(3.93e-8,2.38e-8,1.44e-8)$ PSNR=$(32.88,32.61,34.58)$. Bottom: relative error.}
\label{fig:ux2_res}
\end{figure}

We repeated the experiment with an additional phantom, see Figure~\ref{fig:sig2}. 
The impedance values associated with the background, ellipses and circle were set so $1$, $5$ and $2$. In this case $\mu=1e-4$, learning rate=$1e-2$ and all other parameters as before. 
The results for both $u$ and $\partial u/\partial x$ are presented in figures~\ref{fig:u2_res} and~\ref{fig:ux2_res} respectively.

\section{Inverse Problem}
\begin{figure}
\begin{center}
\subfigure[$n=1$]{\includegraphics[width=.32\linewidth]{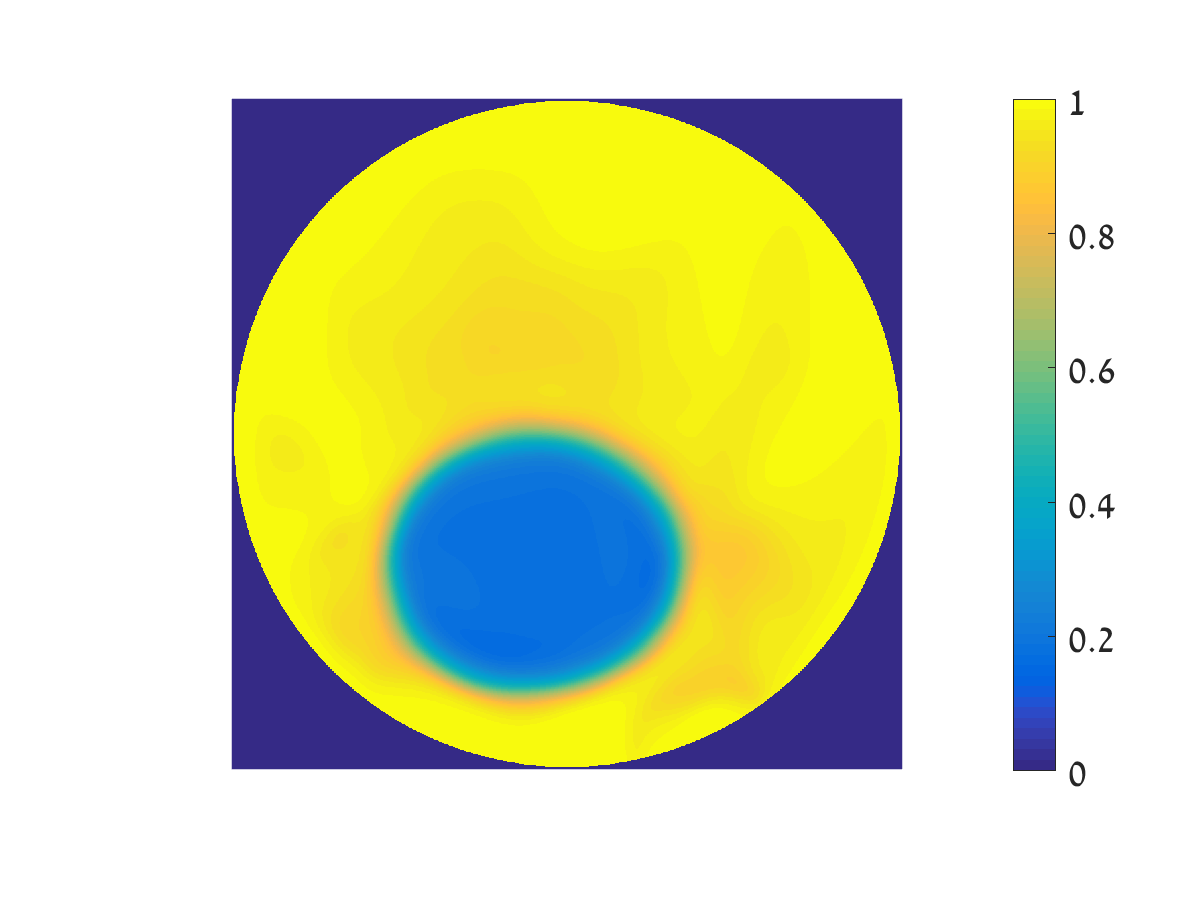}}
\subfigure[$n=2$]{\includegraphics[width=.32\linewidth]{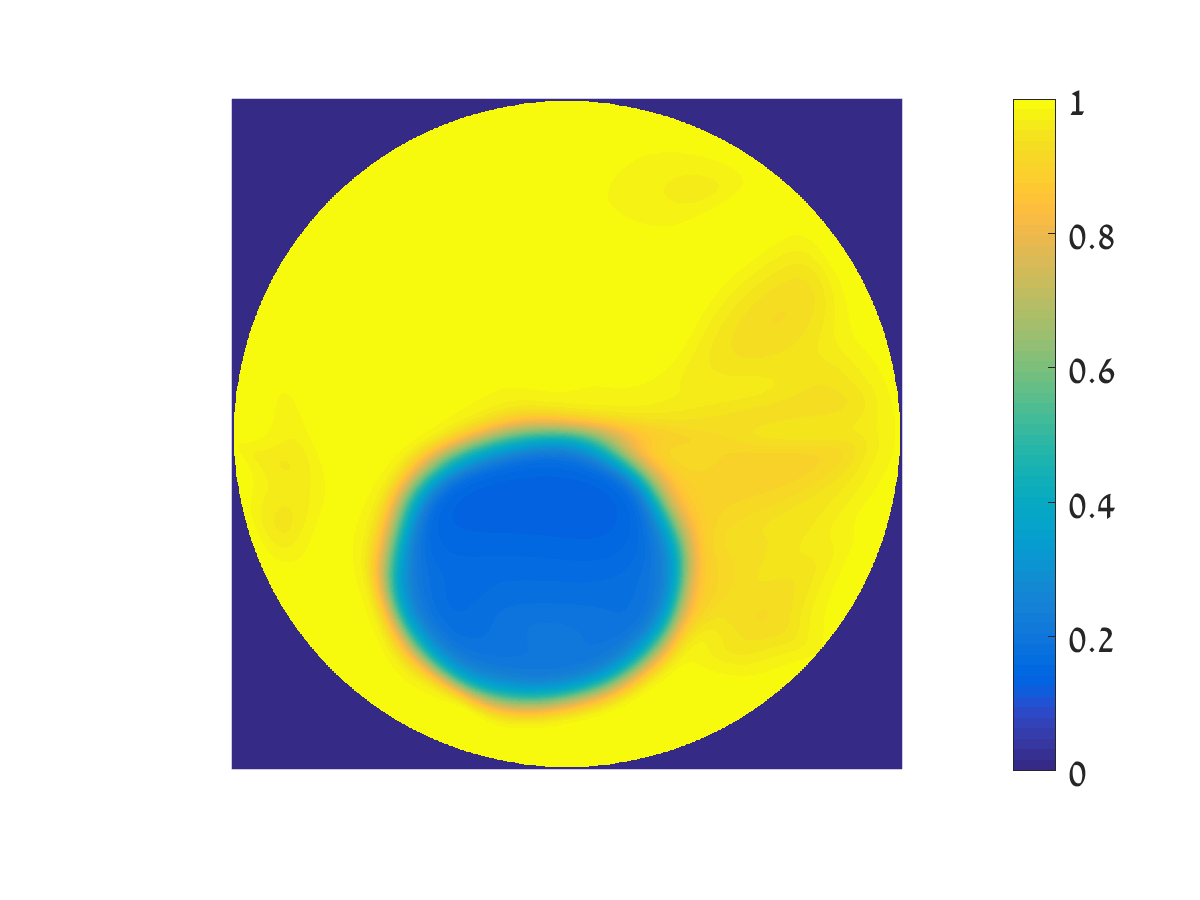}}
\subfigure[$n=3$]{\includegraphics[width=.32\linewidth]{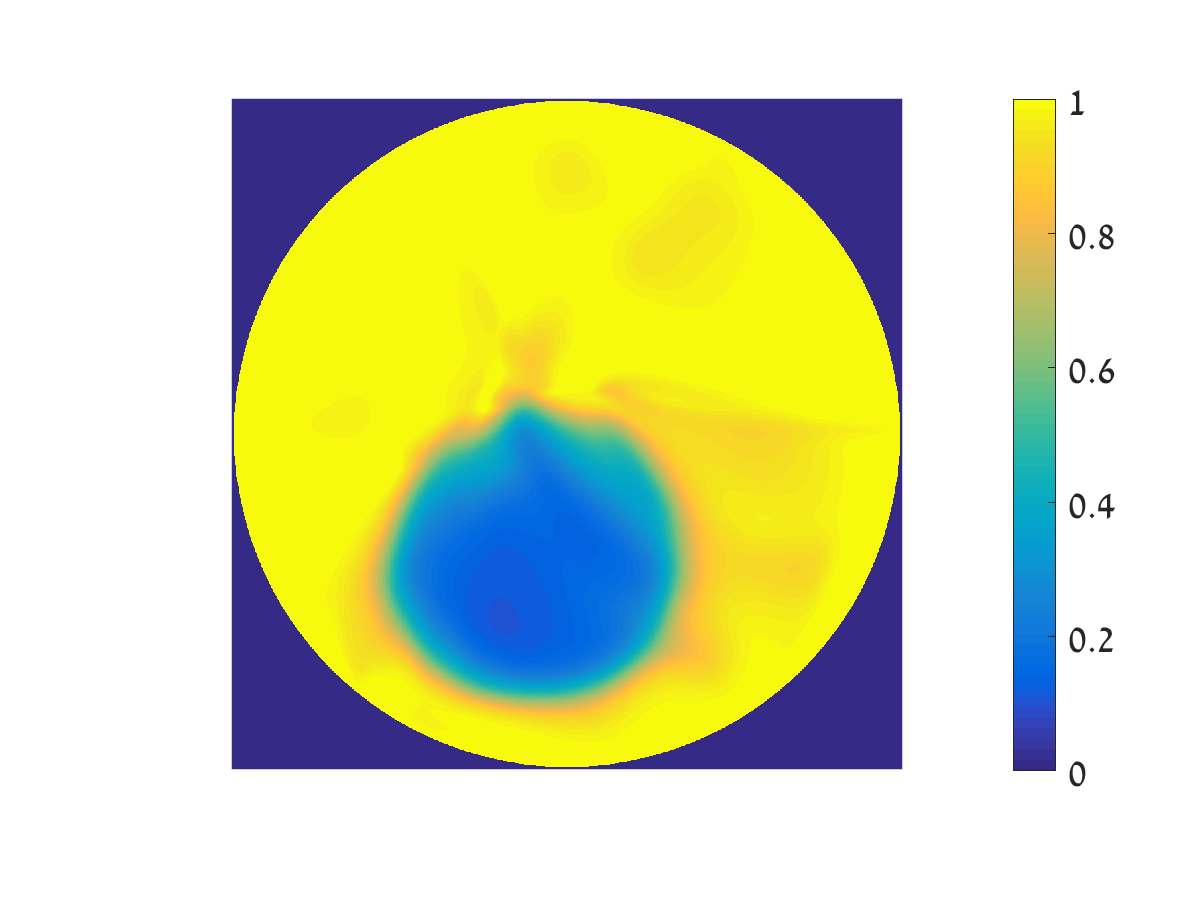}}
\end{center}
\caption{Reconstructed $\sigma$ of phantom $1$. MSE = $(0.22,0.22,0.22)$ PSNR = $(6.45,6.45,6.42)$}
\label{fig:recsig1}
\end{figure}

\begin{figure}
\begin{center}
\subfigure[$n=1$]{\includegraphics[width=.32\linewidth]{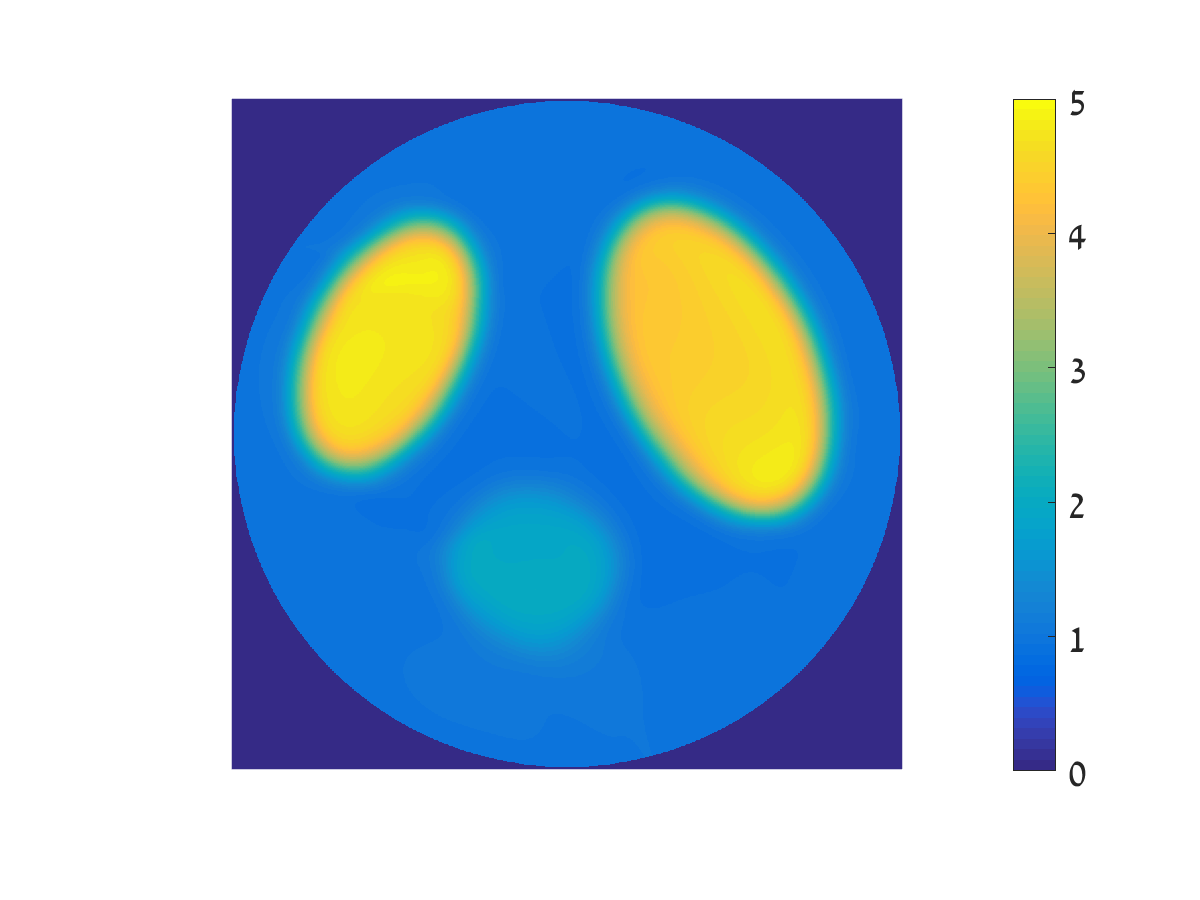}}
\subfigure[$n=2$]{\includegraphics[width=.32\linewidth]{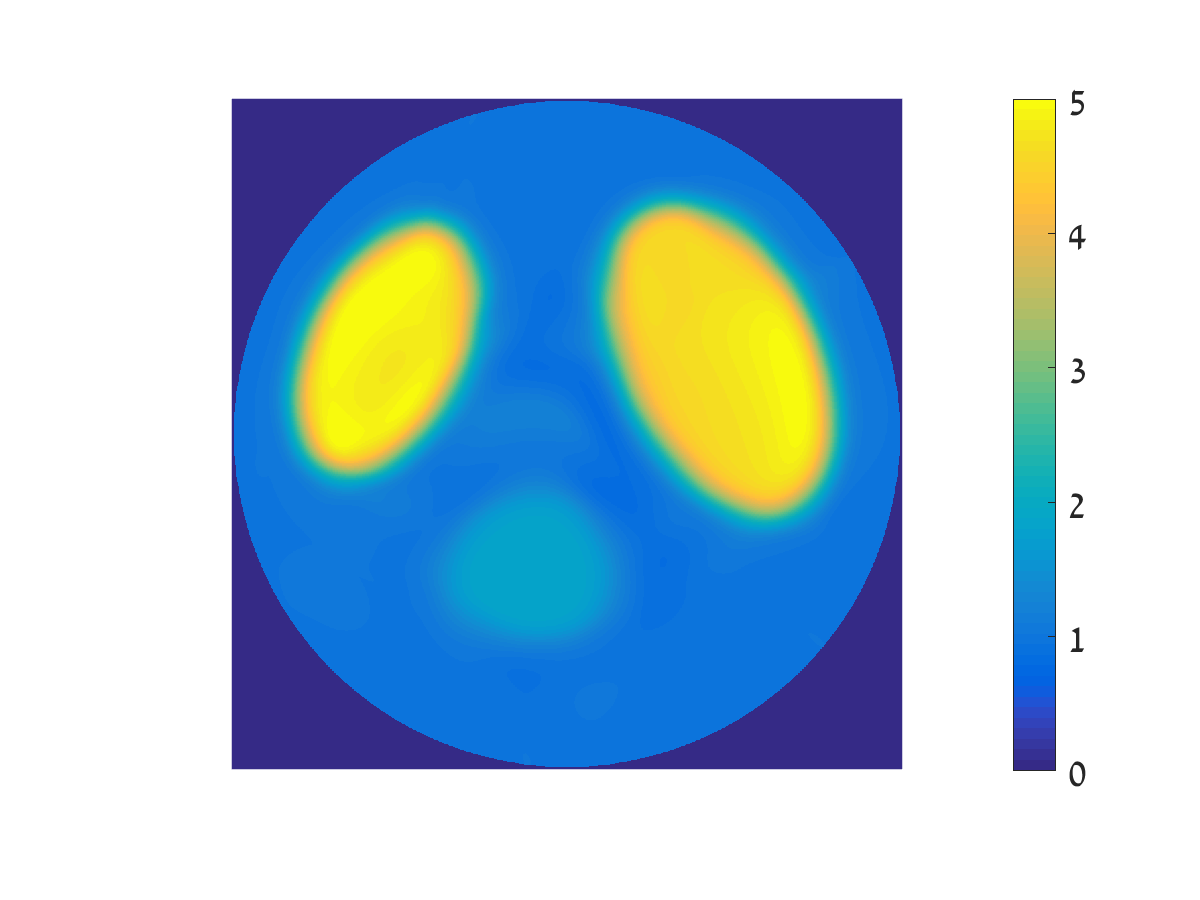}}
\subfigure[$n=3$]{\includegraphics[width=.32\linewidth]{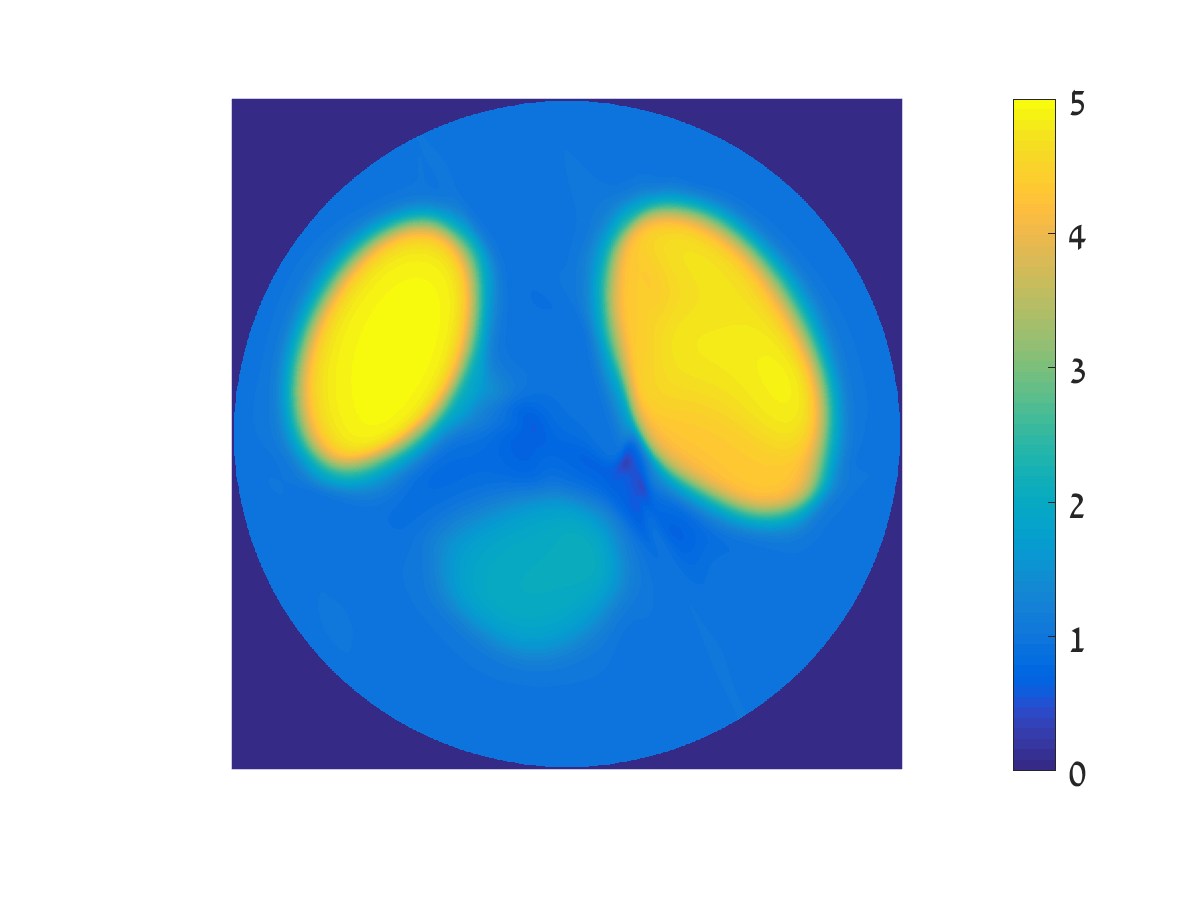}}
\end{center}
\caption{Reconstructed $\sigma$ of phantom $2$. MSE = $(0.26,0.25,0.25)$ PSNR = $(19.80,19.97,19.97)$}
\label{fig:recsig2}
\end{figure}

In the inverse problem, the electrical potential $u(x)$ is known while $\sigma(x)$ is unknown. Since we have a network which approximates $u(x)$, we can evaluate it at any point $x$. The objective function~\eqref{eq:Fi} then takes the form
\begin{equation}
\begin{aligned}
\mathcal{I}(\sigma(x;w_\sigma))= &\frac{\lambda}{N_s}\sum_{i=1}^{N_s} |\mathcal{L}_{i}|^2+\frac{\mu}{K}\sum_{k\in \text{top}_K(|\mathcal{L}_i|)}|\mathcal{L}_k|\\+&\frac{1}{N_b}\sum_{b=1}^{N_b}\Big|\sigma(x_b)-\sigma_0(x_b)\Big|+\alpha\|w_\sigma\|_2^2+\frac{\beta}{N_s}\sum_{i=1}^{N_s}|\nabla\sigma(x_i)|^p.
\end{aligned}
\end{equation}
As in the forward problem, the first two terms enforce $\sigma$ to satisfy the PDE, where $\mathcal{L}_i$ is defined in~\eqref{eq:Li}. The third term imposes the boundary conditions, and the fourth regularizes the network parameters. The last term is the total variation regularization ($p=1$) which promotes the solution towards a piecewise constant solution. 
The network architecture and other parameters are as in the forward problem except for $\beta=1e-3$  and $\mu= 1e-3$. 
Conductivity reconstructions are shown in figures~\ref{fig:recsig1} and~\ref{fig:recsig2} with $\sigma_0=1$ 

\section{Discussion}
Deep networks by their nature use compositions of simple functions such as matrix multiplication and non-linear activations like sigmoid or tanh. 
This structure (i) enables the approximation of an arbitrary function and (ii) is inherently differentiable.
The network architecture dictates the number of degrees of freedom which in turn enables the expressibility of complex functions.
In this work we present a unified framework for the solution of forward and backward problems in partial differential equations. 
The algorithm relies on direct approximation of the unknown function by a neural network which yields an \emph{analytical} smooth solution in a predefined domain. 
The network is trained to satisfy the PDE and boundary conditions in an unsupervised fashion by the minimization of a cost function. The optimization procedure depends on random points set within the domain and its boundary. The problem is therefore mesh free with free-form  domain. We introduce a cost function which is composed of both $L_2$ and $L_\infty$ fidelity terms and additional regularizers. 
The algorithm is demonstrated by an elliptic system in $\mathbb{R}^2$ applied to Electrical Impedance Tomography for both forward and inverse problems. Promising results were achieved for complex and non monotonic functions.   
This framework is general and opens up a wide range of applications and extensions for further research.


\bibliography{pde1}
\bibliographystyle{plain}
\end{document}